\documentclass[conf]{new-aiaa}
\usepackage[utf8]{inputenc}

\usepackage[linesnumbered,ruled,vlined]{algorithm2e}
\usepackage[noend]{algpseudocode}
\usepackage[version=4]{mhchem}
\usepackage{siunitx}
\usepackage{longtable,tabularx}
\usepackage{graphicx,color}
\usepackage{float}
\usepackage{bbm}
\usepackage{caption}
\usepackage{subcaption}
\usepackage{comment,multirow}
\usepackage{tikz}
\usetikzlibrary{shapes,arrows}
\usepackage{bm}
\usepackage{xcolor}

\newcommand{\real}{\mathbb{R}}

\newcommand{\f}{\mathbf{f}}
\newcommand{\x}{\mathbf{x}}

\newcommand{\g}{\mathbf{g}}
\newcommand{\h}{\mathbf{h}}
\newcommand{\boldx}{\mathbf{x}}

\title{Regularized Infill Criteria for Multi-objective Bayesian
Optimization with Application to Aircraft Design}

\author{R. Grapin\footnote{Master student at ISAE-SUPAERO, robingrapin@orange.fr},}

\affil[]{ISAE-SUPAERO, Université de Toulouse, 31055 Toulouse, FRANCE
}

\author{Y. Diouane\footnote{Professor, Mathematics and Industrial Engineering Department, youssef.diouane@polymtl.ca},}

\affil[]{Polytechnique Montréal, Montr\'eal, QC, Canada.
}

\author{J. Morlier\footnote{Professor, Structural Mechanics Department, joseph.morlier@isae-supaero.fr, AIAA Member.},}
\affil{ICA, Universit\'e de Toulouse, ISAE-SUPAERO, MINES ALBI, UPS, INSA, CNRS, Toulouse, France}

\author{N. Bartoli\footnote{Senior researcher, Information Processing and Systems Department, nathalie.bartoli@onera.fr, AIAA MDO TC Member.}, T. Lefebvre\footnote{Research Engineer, Information Processing and Systems 
Department, thierry.lefebvre@onera.fr, AIAA Member.}, 
P. Saves\footnote{PhD Student, Information Processing and Systems Department \& Complex Systems and Engineering Department, paul.saves@onera.fr}, AIAA Member.}
\affil{ONERA,  Universit\'e de Toulouse, Toulouse, France}

\author{J.H. Bussemaker\footnote{Researcher, MDO group, Aircraft Design \& System Integration, Hamburg, jasper.bussemaker@dlr.de, AIAA Member.} }
\affil{DLR, Institute of System Architectures in Aeronautics, Hamburg, Germany}

\begin{document}

\maketitle
\begin{abstract}
Bayesian optimization is an advanced tool to perform efficient global optimization. It consists on  enriching iteratively surrogate Kriging models of the objective and the constraints (both supposed to be computationally expensive) of the targeted optimization problem. Nowadays, efficient extensions of Bayesian optimization to solve expensive multi-objective problems are of high interest. The proposed method, in this paper, extends the super efficient global optimization with mixture of experts (SEGOMOE) to solve constrained multi-objective problems.  To cope with the ill-posedness of the multi-objective infill criteria, different enrichment procedures using regularization techniques are proposed.
The merit of the proposed approaches is shown on known multi-objective benchmark problems with and without constraints.
The proposed methods are then used to solve a bi-objective application related to conceptual aircraft design  with five unknown design variables and three nonlinear inequality constraints. 
The preliminary results show a reduction of the total cost in terms of function evaluations by a factor of 20 
compared to the evolutionary algorithm NSGA-II.
\end{abstract}
\section{Introduction}
\label{section 1}

\lettrine{I}n this work, we are interested in the following continuous constrained multi-objective problem:
\begin{eqnarray} \label{pr:MOO}
& \displaystyle  \min_{x \in \Omega} & \left\{ \f(x) :=[f_1(x),f_2(x),\ldots,f_n(x)]^{\top} ~~ \mbox{subject to}~
~\g(x)\ge 0 ~\mbox{and}~\h(x)=0 \right\} \end{eqnarray}
where $\f:\real^d \to \real^n$ is a vector objective function, $\g: \real^d \to \real^p$ returns the inequality constraints and $\h: \real^d \to \real^m$ the equality constraints. The design space $\Omega \subset \real^d$ is a bounded domain. \\
Many practical optimization problems are of the type~(\ref{pr:MOO}) with usually several conflicting objectives and no solution optimizing all objective functions simultaneously exists in general. Therefore, one needs to select a final solution among Pareto optimal solutions taking into account the balance among objective functions, which is called ``trade-off analysis''. Given by the set of nondominated solutions, being chosen as optimal such that no component of the $\f$ can be improved without sacrificing at least one other component. On the other hand a solution $x^*$ is referred to as Pareto-dominated by another solution $x$ if and only if $x$ is equally good or better than $x^*$ with respect to all the components of $\f$.\\
Motivated by aircraft design problems, we consider in this work that the objective $\f$ and the constraint functions ($\g$ and $\h$) are expensive-to-evaluate  and given in black box form. Typical examples can be found in Multidisciplinary Design Optimization (MDO) problems~\cite{RaymerAircraftDesignConceptual2018,BRAC_AIAA:20} where the objective and the constraints are formulated by several  different disciplines involved (e.g., structure, aerodynamic and propulsion in aircraft design) and their interconnections. In this context, Bayesian optimization (BO)~\cite{MockusBayesianmethodsseeking1975} is a powerful strategy for solving problem~(\ref{pr:MOO}).

In the context of mono-objective optimization, BO consists on modeling the objective and constraint functions with surrogate models, cheap to evaluate, 
built using an initial set of evaluations, called Design of Experiments (DoE)~\cite{MockusBayesianmethodsseeking1975,JonesEfficientglobaloptimization1998}. These models are then minimized and updated iteratively with new evaluations of the problem at wisely chosen points. In fact, by maximizing an acquisition function over the approximate feasible domain~\cite{JonesEfficientglobaloptimization1998}, one is able to improve the knowledge of the objective function over the feasible domain. The maximizer is then used to enrich the model around the assumed optima. The same process is repeated until we reach a maximum number of iterations. The super efficient global optimization with mixture of experts (SEGOMOE) framework~\cite{wb2s,bettebghor2011surrogate,priem2019use,priem2020upper} is an extension of the well-known unconstrained efficient global optimization framework ~\cite{JonesEfficientglobaloptimization1998} to handle  constrained single-objective optimization problems.

For multi-objective Bayesian optimization problems, scalarization techniques have been widely used in the literature~\cite{FeliotBayesianapproachconstrained2017,Knowles_2006}. These techniques transform the multi-objective problem into a single-objective one and thus the related acquisition functions for the transformed mono-objective problem turn to be sub-optimal. By using such approach, obtaining few optimal points may be easy, but constructing the whole Pareto front (particularly, in the case of expensive-to-evaluate functions) may be out of reach. Natural extensions of  mono-objective acquisition functions to the multi-objective case have been an active research area~\cite{Zitzler_etal_2003,Wagner_etal_2010, Ponweiser_etal_2008,Emmerich_etal_2006,Rahat_etal_2017,Zuluaga_etal_2013,Zuluaga_etal_2016}. For example, the expected improvement ($EI$)~\cite{JonesEfficientglobaloptimization1998} has been extended to  the expected hyper-volume improvement ($EHVI$)~\cite{Zitzler_etal_2003,Emmerich_etal_2006}, the probability of improvement ($PI$)~\cite{Jones_2001} to the minimum of probability of improvement ($MPI$)~\cite{Rahat_etal_2017}, the upper confidence bound ($UCB$) to the Pareto Active Learning ($PAL$)~\cite{Zuluaga_etal_2013,Zuluaga_etal_2016},  the stepwise uncertainty reduction ($SUR$)~\cite{Picheny_2014} method has been also extended to multi-objective optimization~\cite{Picheny_2015}. 

In this work, we propose to adapt the SEGOMOE approach to solve constrained multi-objective optimization problems. 
To cope with the ill-posedness of the multi-objective infill criteria, different enrichment procedures using regularization techniques are proposed within SEGOMOE. The proposed regularization technique can be seen as a natural extension of the Watson and Barnes infill criteria ($WB2$~\cite{watson1995infill}  and the scaled $WB2$~\cite{wb2s}) to the multi-objective setting.
Inspired by the existing literature, different acquisition functions will be tested. The potential of the proposed method will be shown on a set of unconstrained and constrained multi-objective problems.
Finally, we present a bi-objective application related to conceptual aircraft configuration related to the CEntral Reference Aircraft System (“CERAS”) and based on the Airbus A320 aircraft data. Upon this test case, we will show the superiority of our proposed approach.

This paper is organized as follows. In Section~\ref{section 2}, the single-objective Bayesian optimization and the SEGOMOE framework are presented. The adaptation of SEGOMOE to the constrained multi-objective context is detailed in Section~\ref{section 3}. The proposed method is then validated in Section~\ref{section 4} on several well known benchmark problems (with and without constraints).  The aircraft design test case is commented in Section~\ref{section 5}.  Conclusions and perspectives are finally drawn in Section~\ref{section 6}.


\section{Single-objective Bayesian Optimization \& SEGOMOE}
\label{section 2}

This section introduces the constrained Bayesian optimization framework~\cite{MockusBayesianmethodsseeking1975,JonesEfficientglobaloptimization1998} that aims to solve the mono-objective optimization problem~\eqref{pr:MOO} (here $n=1$) with a minimal number of calls.
To do so,  Gaussian Process (GP) (also known as Kriging)~\cite{RasmussenGaussianprocessesmachine2006,Krigestatisticalapproachbasic1951} are trained from a design of experiments (DoE) (i.e. set of designs evaluated on the objective and constraint functions) of $l$ points. GPs are then used to provide, with a cheap computational cost, a prediction $\mu_{s}^{(l)}: \mathbb{R}^d \mapsto \mathbb{R}$ and an associated uncertainty $\sigma_{s}^{(l)}: \mathbb{R}^d \mapsto \mathbb{R}$ for each point of $\bm{x} \in \Omega$ where $s : \mathbb{R}^d \mapsto \mathbb{R}$ can be either $f$, $g_i$ for a given inequality constraint component $i$ ($\forall i\in[1,p]$) or  $h_j$ for a given equality constraint component $j$ ($\forall j\in[1,m]$). \\
Concerning the objective function, these information are combined in an acquisition function  $\alpha_f^{(l)}: \mathbb{R}^d \mapsto \mathbb{R}$~\cite{frazier2018tutorial,Bartoliadaptivemodeling2019,WangMaxvalueentropysearch2017} coding the trade-off between exploration of the highly uncertain domain that can hide a minimum and exploitation of the minimum of the GP prediction. 
For the constraints, these information are joined to produce two feasibility criteria $\alpha_{\bm{g}}^{(l)}: \mathbb{R}^d \mapsto \mathbb{R}^p$ and $\alpha_{\bm{h}}^{(l)}: \mathbb{R}^d \mapsto \mathbb{R}^m$~\cite{frazier2018tutorial,lam2015multifidelity,priem2019use} which are generally explicit.
The point $\bm{x}^{(l+1)}$, solving the constrained maximization trade-off sub-problem: 
\begin{equation}
    \bm{x}^{(l+1)}=\arg\max\limits_{\bm{x} \in \Omega} \left\{ \alpha_f^{(l)}(\bm{x}) ~~\mbox{s.t.}~~ \bm{x}\in \Omega_{\bm{g}}^{(l)} \displaystyle \cap \Omega_{\bm{h}}^{(l)} \right\},
    \label{eq:vst_prob}
\end{equation}
where $\Omega_{\bm{g}}^{(l)}$ (resp. $\Omega_{\bm{h}}^{(l)}$) is the approximate feasible domain defined by the feasibility criterion $\bm{\alpha}_{\bm{g}}^{(l)}$ (resp. $\bm{\alpha}_{\bm{h}}^{(l)}$), is thus iteratively added to the DoE until a maximum number of iterations $max\_nb\_it$ is reached.
The solution provided to the problem~\eqref{pr:MOO} is eventually the best point in the DoE (i.e with the minimal feasible value of $f_1$). 

In order to handle constrained optimization problems with a large number of design variables, a general framework called SEGOMOE for Super Efficient Global Optimization coupled with Mixture Of Experts
has been proposed by ONERA \& ISAE-SUPAERO. The initial algorithm Super Efficient Global Optimization
(SEGO) algorithm~\cite{SasenaExplorationmetamodelingsampling2002} has been enhanced by the use of Mixture of Experts (MOE)~\cite{bettebghor2011surrogate}, the Kriging with Partial Least Squares (KPLS)~\cite{BouhlelImprovingkrigingsurrogates2016} for high dimensional problems, different acquisition function criteria for highly non linear objective functions~\cite{wb2s}  or constraint functions~\cite{priem2020upper}. The kriging based models and the mixture of experts are built using the opensource toolbox SMT\footnote{\url{https://github.com/SMTorg/smt}}~\cite{smt}.
Finally, the search of the optimum is done using different optimizers capable of
considering non linear constraints based either on derivative free optimizer (such as COBYLA -Constrained
Optimization BY Linear Approximation~\cite{cobyla}) or based on gradient method as SLSQP (for Sequential Least
Squares Programming~\cite{slsqp}) or SNOPT (Sparse Nonlinear OPTimizer~\cite{snopt}) both using the Jacobian calculation of the mixture of experts (for the objective and the
constraint function GP analytic derivatives are used). \\
The resulting toolbox named the super efficient global optimization with mixture of experts (SEGOMOE) has been validated on different analytical and industrial test cases~\cite{SEGOMOE_ISSMO:16, SEGOMOE_AIAA:17, AGILE_Nacelle_OneraTsagi:18, Bartoliadaptivemodeling2019, BRAC_AIAA:20}. Finally, the summary of SEGOMOE algorithm steps is given in Fig.~\ref{SEGO_optim_flowchart}.

\tikzstyle{decision} = [diamond, draw,text width=5em, text badly centered, node distance=3cm, inner sep=0pt]
\tikzstyle{block} = [rectangle, draw,text width=8em, text centered, rounded corners, minimum height=4em]
\tikzstyle{line} = [draw, -latex']
				
\begin{figure}[h!]
					\centering
					\begin{tikzpicture}[node distance = 1.5cm, auto]
						\node [block,text width=15em] (init) {Constructing the initial design of experiments};
						\node [block,text width=7.5em, below of=init,node distance = 2cm] (metamodele) {Re/building metamodels based on GPs};
						\node [block,text width=9.5em, below of=metamodele,node distance = 2.1cm] (optim_critere) {Solving problem ~\eqref{eq:vst_prob} to find $\bm{x}^{(l+1)}$ };
						\node [block,text width=10.5em, below of=optim_critere,node distance = 2.1cm] (eval) {Computing the enrichment point: evaluate the objective and constraint functions at $\bm{x}^{(l+1)}$};
						\node [block, below of=eval,text width=7.5em,node distance = 2.1cm] (decide_optim) {Stopping criterion: $max\_nb\_it$\; reached?};
						\node [block, left of=decide_optim,node distance=6cm,text width=10em] (fin) {Stop and return the best point found in the DoE};
						\node [block, text width=7.5em,right of=optim_critere,node distance=3.5cm] (reconstruction) {Updating the design of experiments};
						\path [line] (init) -- (metamodele);
						\path [line] (metamodele) -- (optim_critere);
						\path [line] (optim_critere) -- (eval);
						\path [line] (eval) -- (decide_optim);
						\path [line] (decide_optim) -- node [anchor=north]{Yes} (fin);
						\path [line] (decide_optim) -| node [anchor=north]{No} (reconstruction);
						\path [line] (reconstruction) |- (metamodele);
					\end{tikzpicture}
					\caption{\label{SEGO_optim_flowchart} The summary of SEGOMOE algorithm steps.}
				\end{figure}
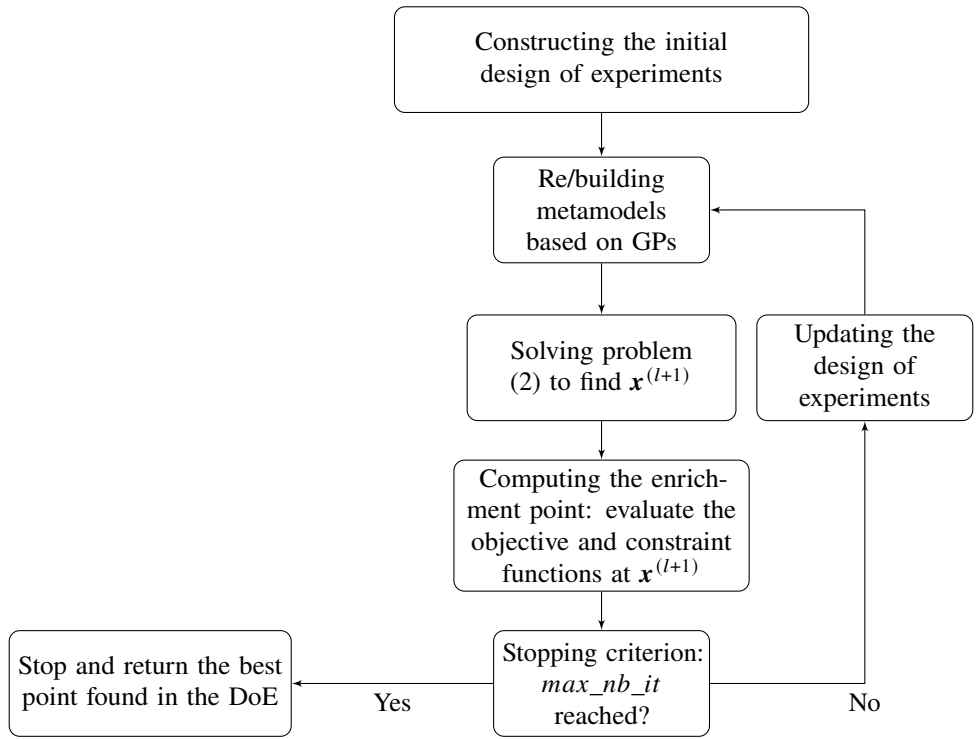

\section{Regularized infill criteria for multi-objective Bayesian
optimization within SEGOMOE}
\label{section 3}

\subsection{Extension of SEGOMOE to solve constrained multi-objective problems}
In this section we present how the SEGOMOE solver can be extended to solve constrained multi-objective problems of the form~\eqref{pr:MOO}. The SEGOMOE is modified such that one can get an approximation of the optimal Pareto front related to problem~\eqref{pr:MOO}. For that sake, iteratively, we will apply the same procedure as SEGOMOE (see Fig.~\ref{SEGO_optim_flowchart}) except three changes.\\ 
The first change consists in the construction of a GP surrogate of each component of the objective function $\f$. We note that this step can be executed in parallel for all the components of $\f$, $\h$ and $\g$.\\
The second change introduced in SEGOMOE is related to the construction of a new scalar acquisition function that takes into account the multi-objective nature of the problem. At the stage, we work with existing multi-objective functions (such as $EHVI$, $MPI$, $PI$, ...), noted $\alpha_{\f}$. The maximization of such acquisition function within the approximate feasible domain $\Omega_{\g}\cap \Omega_{\h}$ will allow the user to enrich the DoE adequately. The same procedure is applied iteratively until a maximum number of iterations is reached. \\
The third change comes at the end of the iterative process. In fact, using the updated GP surrogates of the objectives and the constraints, we will build an approximate Pareto front using a classical constrained multi-objective solver. For instance, in our experiments, we used the well-known genetic algorithm NSGA-II solver~\cite{nsga2}. The latter optimization is computationally inexpensive as it applies only to GP surrogates.\\
The complete description of our proposed extension of SEGOMOE to handle constrained multi-objective problems is given in Algorithm~\ref{MO:BO:algo}.

\begin{algorithm}
\caption{SEGOMOE for constrained  multi-objective Bayesian optimization.}\label{MO:BO:algo}
\KwData{Objectives $\f$ and constraint functions ($\h$ and $\g$), $\Omega$ the design space, the initial DoE for objectives and constraints, a maximum number of iterations (\textit{max\_nb\_it}) \;}
\For{l=0 \textbf{to} max\_nb\_it}{
Re/Building the GP surrogates for the objectives and the constraints\;
Using the GP surrogates, build $\Omega_{\bm{g}}^{(l)}$ and $\Omega_{\bm{h}}^{(l)}$ the approximate feasible domains related to $\g$ and $\h$\;
Selecting a scalar acquisition function $\alpha^{(l)}_{\f}$ (e.g., $EHVI$, $PI$, $MPI$, $PAL$, $SUR$, ...)\;
Solving the enrichment problem $ \bm{x}^{(l+1)}=\arg\max\limits_{\bm{x} \in \Omega} \left\{ \alpha_{\f}^{(l)}(\bm{x}) ~~\mbox{s.t.}~~ \bm{x}\in \Omega_{\bm{g}}^{(l)} \displaystyle \cap \Omega_{\bm{h}}^{(l)} \right\}$\;
 Evaluating the objectives and the constraints at the new point $\bm{x}^{(l+1)}$ and updating the DoE\;
}
 Building an approximate Pareto front by apply a constrained multi-objective solver (e.g., a genetic algorithm) on the GP surrogates of the objectives and the constraints\;
\KwResult{An approximation of the optimal Pareto front.}
\end{algorithm}

\subsection{Regularized infill criteria for SEGOMOE}
\label{section 3reg}
During the enrichment process, the maximization of the infill criterion might not be an easy task. In fact, the large number of objectives, the dimension of the design space, the presence of noisy on the computation make  enrichment process very ill-posed (i.e., badly conditionned, many local maxima, sensitive to noisy, ...). In this context, using regularization techniques to include additional information can be very useful for an efficient enrichment step. In mono-objective optimization, regularization techniques are known to lead a significant improvement within BO.  The Watson and Barnes ($WB2$)~\cite{watson1995infill}  and the scaled $WB2$~\cite{wb2s} criteria are among existing regularization techniques in the literature. In this paper, inspired by the scaled $WB2$ ~\cite{wb2s}, we proposed to extend the use of regularization techniques of the acquisition functions to cover the multi-objective setting as well. For a given multi-objective infill criterion $\alpha_{\f}$, we set the regularized  infill criterion within SEGOMOE as follows
\begin{eqnarray} \label{eq:alpha:reg}
\alpha^{\mbox{reg}}_{\f}(\bm{x}) &=& \gamma\;\alpha_{\f}(\bm{x}) - \psi(\mu_{\f}(\bm{x})), 
\end{eqnarray}
where $\gamma$ is a constant parameter and $\mu_{\f}: \real^d \to \real^n$ is the GP prediction associated with $\f$. The function $\psi: \real^n \to \real$ is a scalarization operator. Different choices exist for the function $\psi$, in the context of this paper, we will investigate two options. Namely, for a given $y\in \real^n$, we consider
\begin{eqnarray*}
(\mbox{reg}=\max)&: & \psi(y) =  \max_{i \leq n}y_i,  \label{eq:max}\\
(\mbox{reg}=\mbox{sum}) & :& \psi(y)= \sum_{i=1}^n y_i. \label{eq:sum}
\end{eqnarray*}
The constant parameter $\gamma$ in \eqref{eq:alpha:reg} is estimated in the spirit of what is done for the scaled $WB2$ infill criterion~\cite{wb2s}, i.e.,
\begin{align*} 
\begin{split}
\; \gamma = \left\{
    \begin{array}{ll}
        \beta \times \frac{\psi(\mu(\bm{x}_{\alpha_{\f}}^{\max}))}{\alpha(x_{\alpha_{\f}}^{\max})}\; &\mbox{ if} \; \alpha(x_{\alpha_{\f}}^{\max}) > 0,\\
        1 &\;\mbox{ otherwise, } 
    \end{array}
\right.
\end{split}
\end{align*}
with $\beta$ is an additional hyperparameter (set to $100$ in~\cite{wb2s}), the vector  $x_{\alpha_{\f}}^{\max}$ is an approximate maximizer of the acquisition function over the feasible domain, i.e., $$x_{\alpha_{\f}}^{\max} \approx \displaystyle \arg \max_{\bm{x} \in \real^d} \alpha_{\f}(\bm{x})   ~~~~\mbox{subject to}~~\bm{x}\in \Omega_{\bm{g}} \displaystyle \cap \Omega_{\bm{h}}.$$

\section{Numerical Results}
\label{section 4}

In this section, the potential of the proposed method is evaluated using $12$ analytical and  challenging multi-objective problems with and without constraints.
\subsection{Benchmark problems}

The chosen test suite is constituted of three unconstrained and three constrained problems.  
The first group of unconstrained problems is composed  of the three problems known as ZDT problems~\cite{ZDT}. Those problems are bi-objective with $d$ design variables, $d\ge 2$.. A detailed description of the three problems is given in Appendix~\ref{appendix1}.  In this paper, we have tested $d=2$, $d=5$ and $d=10$, which lead to a total of $9$ unconstrained problems. The second group of test cases is composed of three constrained multi-objective problems. The first one is the Binh and Korn  problem (BNH)~\cite{binh1997mobes} with 2 objectives,  2 variables and 2 inequality constraints. The second problem is the Tanaka one (TNK)~\cite{zapotecas2018review} with 2 objectives,  2 design variables and 2 inequality  constraints. The third constrained problem is the Osyczka and Kundu problem (OSY)~\cite{osyczka1995new}  with 2 objectives,  6 design variables and 6 inequality constraints. 
The three test problems, detailed in Appendix~\ref{appendix2}, are well known in genetic algorithm community.

\subsection{Implementation details}

The GP surrogate models of the objectives and constraints are done using the SMT opensource toolbox \footnote{\url{https://github.com/SMTorg/smt}}~\cite{smt}. If no initial known points of the problem are provided, the Latin Hypercube Sampling (LHS)~\cite{lhs} sampling method is called to fill equally the whole design space. \\
To obtain the optimal points once the model is enough refined (step 7 in Algo.~\ref{MO:BO:algo}), the multi-objective framework Pymoo~\cite{pymoo} has been chosen. It allows mixed-integer variables and constraints management. The GP mean predictions from the objective and the constraint functions are used to define the problem on which the genetic algorithm is run. These GP surrogate models of the  benchmark test cases are optimized using NSGA-II~\cite{nsga2} from Pymoo~\cite{pymoo} with 50 generations of 100 individuals.

Once the experiment parameters and benchmark problems are set, the comparison of optimization of multi-objective black-box problems can scale the gain in efficiency between the use of different infill criteria. To compare the implemented methods, we fix the budget in function of the design space's dimension. Naming $d$ the dimension of the design space and $c$ the number of constraints, the sampling will be made of $2d+ 2c+1$ points and $20d$ evaluations of the problem will be computed in total, i.e $20d - (2d+2c+1)$ iterations lead to the presented results.

\subsection{Performance indicator}

In the context of multi-objective optimization, the efficiency of the tested methods will be evaluated based on the following two criteria:
\begin{itemize}
    \item the proximity between the obtained not dominated points and the associated explicit Pareto front ones;
    \item the distribution of the obtained points in the objective space to cover as largely as possible the Pareto-optimal possibilities.
\end{itemize}
Different performance indicators exist in the literature. In this context, the property of Pareto compliance is of high interest. Namely, a weakly Pareto compliant indicator $I$ is defined such that for two sets of points $A$ and $B$, if the elements of $A$ dominates those of $B$ then $I(A) \leq I(B)$. One efficient way to estimate the compliance indicator is given by inverted generational distance plus (${IGD}^{+}$)~\cite{ishibuchi2015modified}. The ${IGD}^{+}$ indicator is defined as follows, for a given set of points $A=\left \{a_1,a_2,\ldots,a_{|A|}\right \}$ and a reference set $Z=\left \{z_1,z_2,\ldots,z_{|Z|}\right \}$ with true values from the optimal Pareto front, the inverted generational distance is given by
\begin{equation}
{IGD}^{+}(A) =  \; \frac{1}{|Z|} \; \bigg( \sum_{i=1}^{|Z|} {d_i^{+}}(z_i)^2 \bigg)^{1/2} \;\mbox{ with } \; d_i^{+}(z_i) = \max \{ \min_{a_i \in A}a_i - z_i, 0\} 
\end{equation}
where the notation $|Z|$ defines the cardinal of the set of points $Z$. In our comparison tests, the smaller is the value of ${IGD}^{+}$ the better is the tested method.
We note that evaluating the ${IGD}^{+}$ indicator requires the knowledge of the true optimal Pareto front to build the reference set $Z$. For that reason, we tested analytical multi-objective problems, for which we know explicitly the optimal Pareto fronts.

Due to stochastic effects caused by the starting point at each enrichment step and by the genetic algorithm, the ${IGD}^{+}$ convergence plots are averaged over 10 runs.
\subsection{Obtained results}

\subsubsection{Comparison of existing infill criteria within SEGOMOE}
\begin{figure}[H]
\begin{subfigure}{\textwidth}
\includegraphics[width=0.33\textwidth]{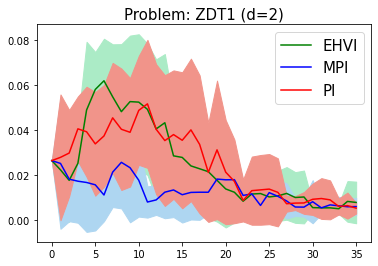}
\includegraphics[width=0.33\textwidth]{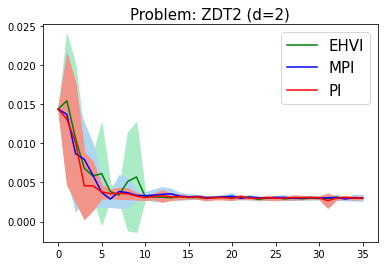}
\includegraphics[width=0.33\textwidth]{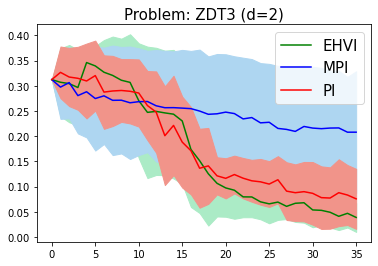}

\includegraphics[width=0.33\textwidth]{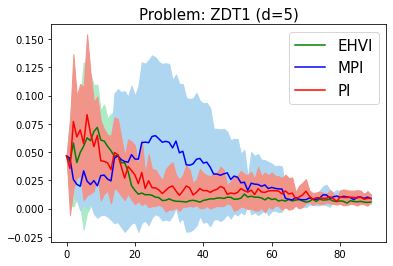}
\includegraphics[width=0.33\textwidth]{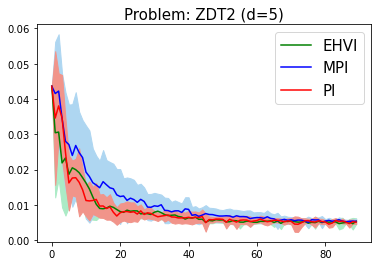}
\includegraphics[width=0.33\textwidth]{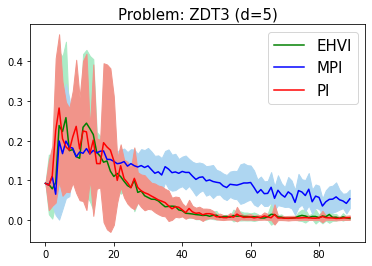}

\includegraphics[width=0.33\textwidth]{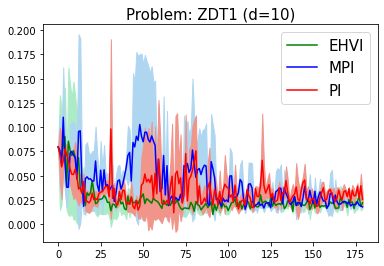}
\includegraphics[width=0.33\textwidth]{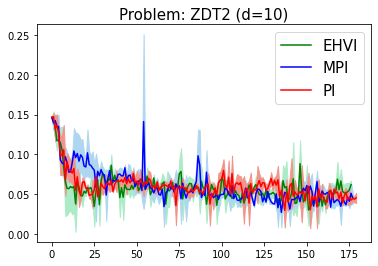}
\includegraphics[width=0.33\textwidth]{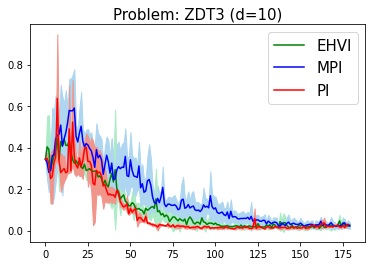}
\caption{Unconstrained problems (from left to right: ZDT1, ZDT2 and ZDT3). }\label{fig:uncons-IGD}
\end{subfigure}

\begin{subfigure}{\textwidth}
\includegraphics[width=0.33\textwidth]{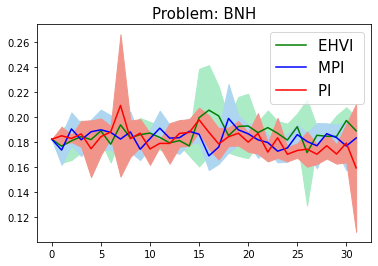}
\includegraphics[width=0.33\textwidth]{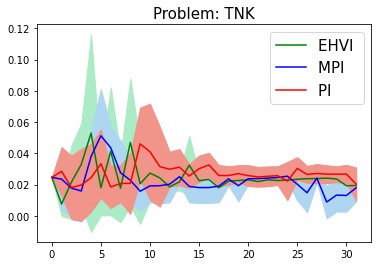}
\includegraphics[width=0.33\textwidth]{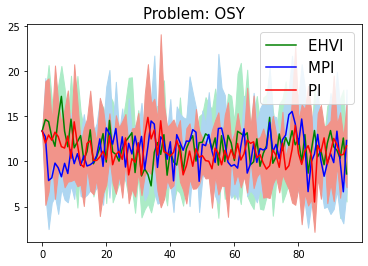}
\caption{Constrained problems (from left to right: BNH, TNK and OSY) }\label{fig:cons-IGD}
\end{subfigure}

\caption{Obtained convergence plots (i.e., ${IGD}^{+}$ values across iterations): a comparison of the acquisition functions $EHVI$, $PI$ and $MPI$ within SEGOMOE using ${IGD}^{+}$.  }\label{fig:IGD}
\end{figure}

\begin{figure}[H]
\begin{subfigure}{\textwidth}
\includegraphics[width=0.33\textwidth]{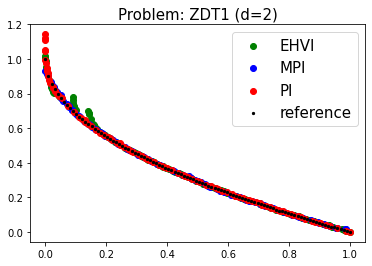}
\includegraphics[width=0.33\textwidth]{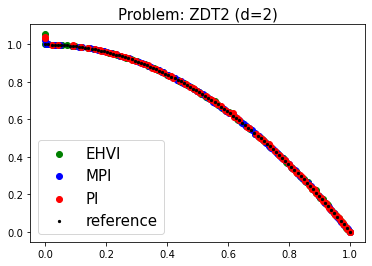}
\includegraphics[width=0.33\textwidth]{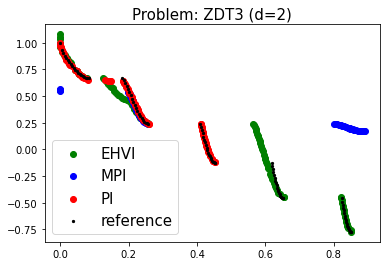}

\includegraphics[width=0.33\textwidth]{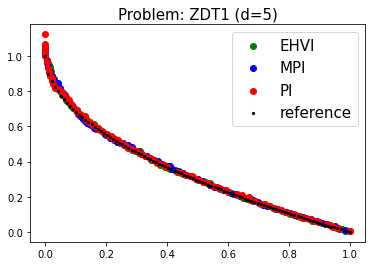}
\includegraphics[width=0.33\textwidth]{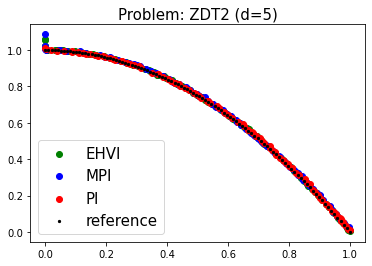}
\includegraphics[width=0.33\textwidth]{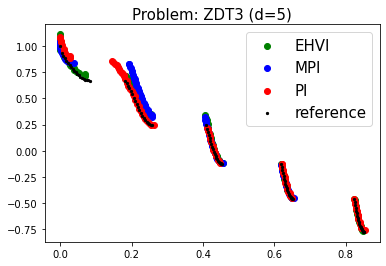}

\includegraphics[width=0.33\textwidth]{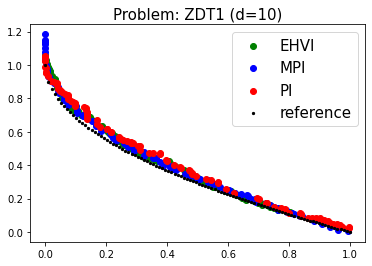}
\includegraphics[width=0.33\textwidth]{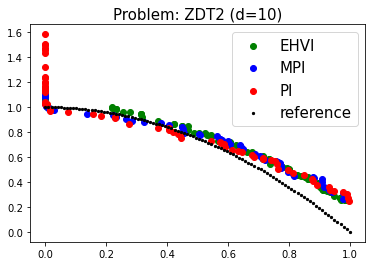}
\includegraphics[width=0.33\textwidth]{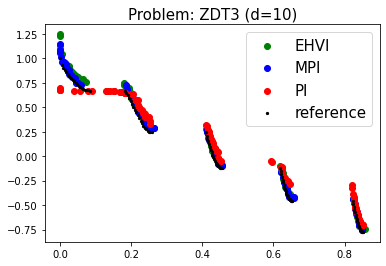}

\caption{ Unconstrained problems (from left to right: ZDT1, ZDT2 and ZDT3). }\label{fig:uncons-pf}
\end{subfigure}

\begin{subfigure}{\textwidth}
  \centering
\includegraphics[width=0.33\textwidth]{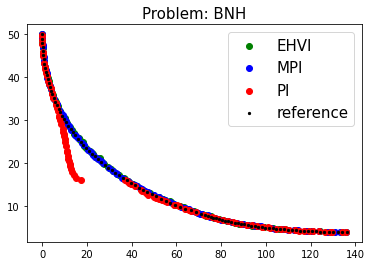} 
\includegraphics[width=0.33\textwidth]{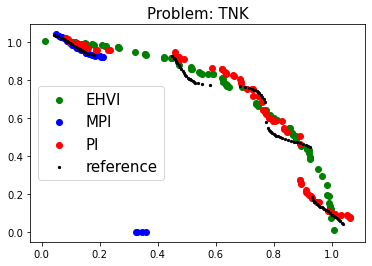}
\includegraphics[width=0.33\textwidth]{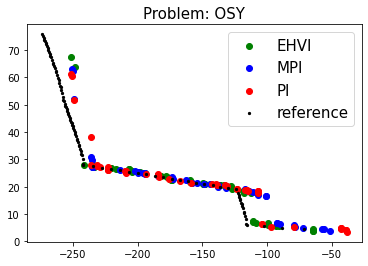}
\caption{Constrained problems (from left to right: BNH, TNK and OSY).}
  \label{fig:cons-pf}
\end{subfigure}
\caption{A comparison of the obtained Pareto fronts obtained using $20d$ points.}  \label{fig:pf}
\end{figure}

Figure~\ref{fig:IGD} illustrates the convergence of the ${IGD}^{+}$ indicator through the iterations for the tested problems. For each problem, different acquisition criteria are compared: $EHVI$, $PI$ and $MPI$ whose mean value and associated dispersion are displayed. The associated Pareto fronts obtained at the final iteration are given in Fig.~\ref{fig:pf}. The  true Pareto front (which is known for those test cases) is also included in the Figure. 

For unconstrained problems, see Fig.~\ref{fig:uncons-IGD}, one can see that SEGOMOE performs well for all the three tested acquisition functions, with comparable convergence plot for ${IGD}^{+}$. $EHVI$ is slightly better compared to $PI$ and $MPI$ for the unconstrained test-cases. For the three constrained problems (BNH, TNK and OSY), convergence plots are given in Fig.~\ref{fig:cons-IGD}. One can see that the problems TNK ans OSY are turn to be very challenging for SEGOMOE compare to the BNH problem. Similarly to the unconstrained test cases, we obtain comparable performance for the acquisition functions $EHVI$, $PI$ and $MPI$.

Last, we note that for all the tested problems for each infill criterion, SEGOMOE converged to a good approximation of the explicit Pareto front. The dispersion over the ${IGD}^{+}$ scores becomes lower over the iterations, showing the consistency of each method. Even if the dimension curse is a phenomenon leading to an exponential increase of the research space, the methods do not lose that much in efficiency, although the number of points calculated is chosen linearly with the problems' dimensions. We can mention than $PI$ and $EHVI$ infill criteria are most of the time the ones giving the best results. The estimation of the $EHVI$ acquisition function is computationally expensive  compared to $PI$ which is  the easiest to compute among the tested infill criteria.

\subsubsection{Regularized infill criteria comparison within SEGOMOE}
\begin{figure}[H]
\begin{subfigure}{\textwidth}
\includegraphics[width=0.33\textwidth]{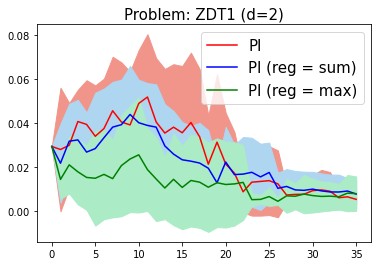}
\includegraphics[width=0.33\textwidth]{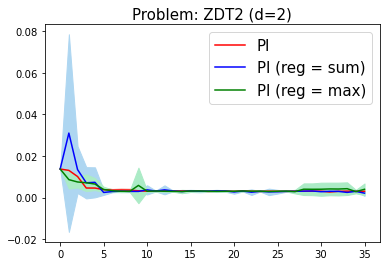}
\includegraphics[width=0.33\textwidth]{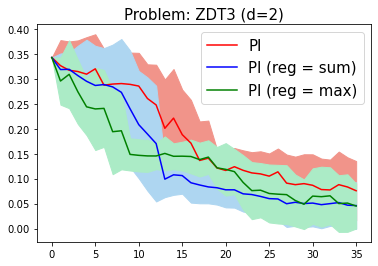}

\includegraphics[width=0.33\textwidth]{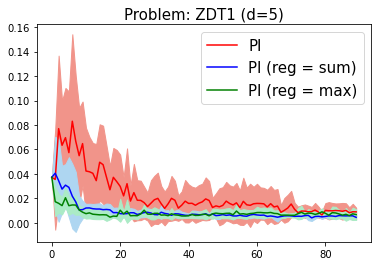}
\includegraphics[width=0.33\textwidth]{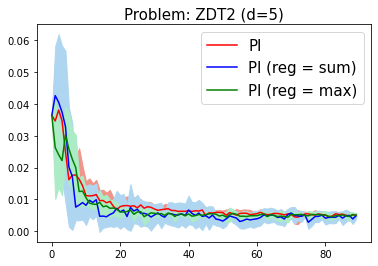}
\includegraphics[width=0.33\textwidth]{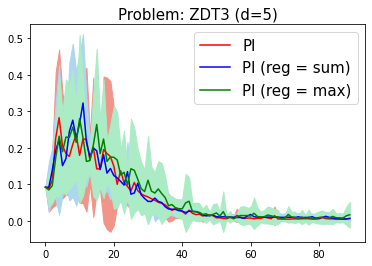}

\includegraphics[width=0.33\textwidth]{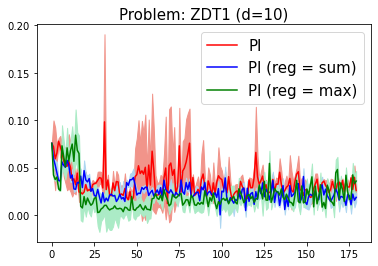}
\includegraphics[width=0.33\textwidth]{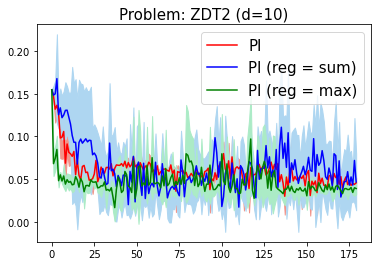}
\includegraphics[width=0.33\textwidth]{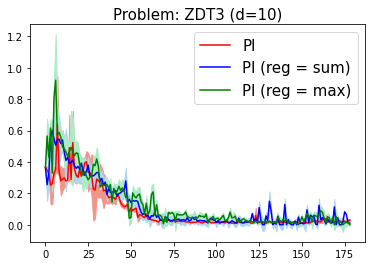}
\caption{Unconstrained problems (from left to right: ZDT1, ZDT2 and ZDT3). }\label{fig:uncons-IGD:reg:PI}
\end{subfigure}

\begin{subfigure}{\textwidth}
\includegraphics[width=0.33\textwidth]{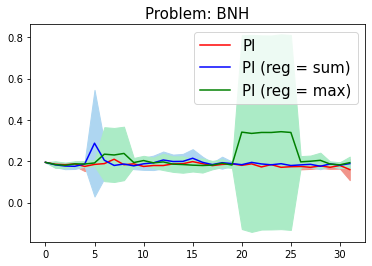}
\includegraphics[width=0.33\textwidth]{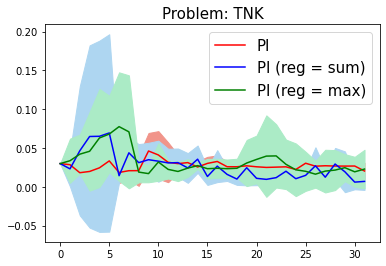}
\includegraphics[width=0.33\textwidth]{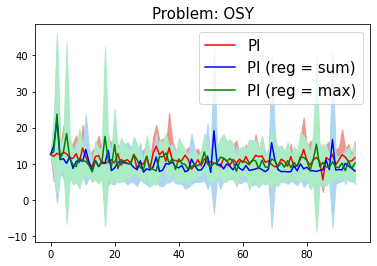}
\caption{Constrained problems (from left to right: BNH, TNK and OSY) }\label{fig:cons-IGD:reg:PI}
\end{subfigure}

\caption{Obtained convergence plots (i.e., ${IGD}^{+}$ values across iterations): regularization effect on the $PI$ acquisition function within SEGOMOE.  }\label{fig:IGD:reg:PI}
\end{figure}

\begin{figure}[H]
\begin{subfigure}{\textwidth}
\includegraphics[width=0.33\textwidth]{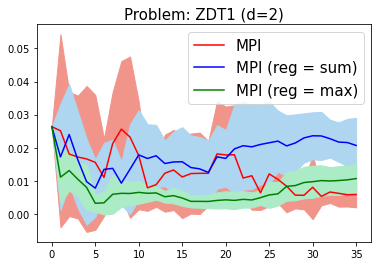}
\includegraphics[width=0.33\textwidth]{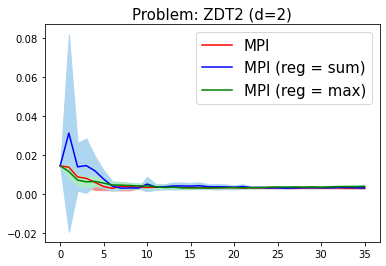}
\includegraphics[width=0.33\textwidth]{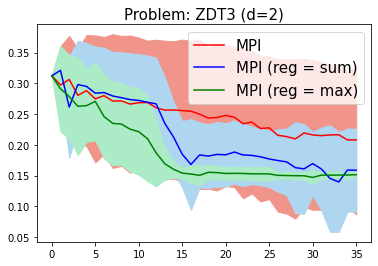}

\includegraphics[width=0.33\textwidth]{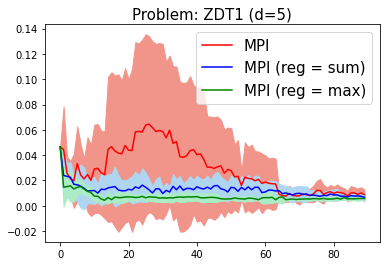}
\includegraphics[width=0.33\textwidth]{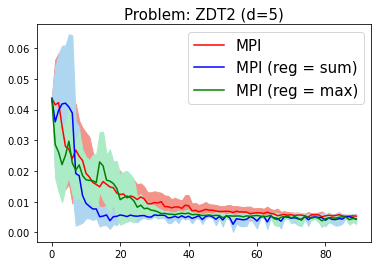}
\includegraphics[width=0.33\textwidth]{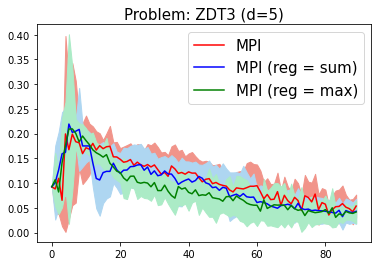}

\includegraphics[width=0.33\textwidth]{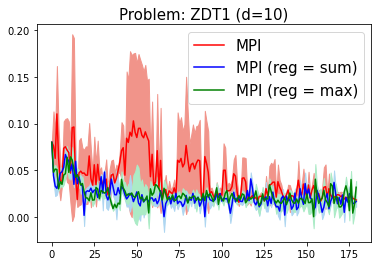}
\includegraphics[width=0.33\textwidth]{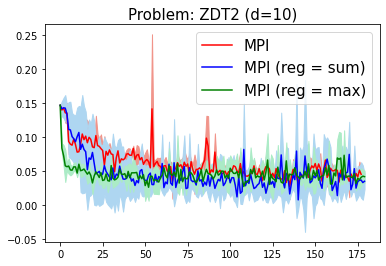}
\includegraphics[width=0.33\textwidth]{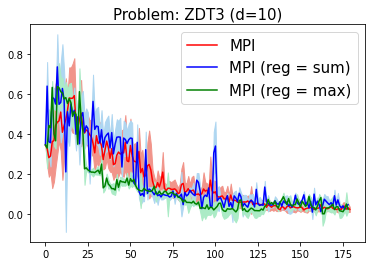}
\caption{Unconstrained problems (from left to right: ZDT1, ZDT2 and ZDT3). }\label{fig:uncons-IGD:reg:MPI}
\end{subfigure}

\begin{subfigure}{\textwidth}
\includegraphics[width=0.33\textwidth]{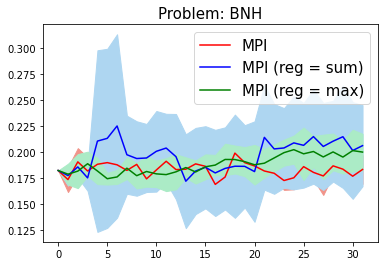}
\includegraphics[width=0.33\textwidth]{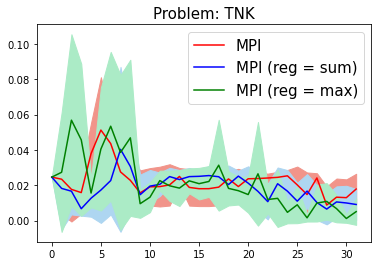}
\includegraphics[width=0.33\textwidth]{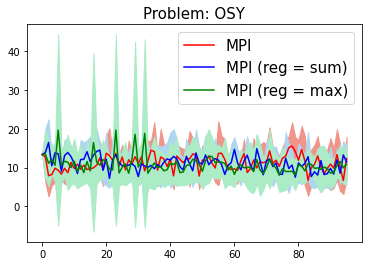}
\caption{Constrained problems (from left to right: BNH, TNK and OSY) }\label{fig:cons-IGD:reg:MPI}
\end{subfigure}

\caption{Obtained convergence plots (i.e., ${IGD}^{+}$ values across iterations): Regularization effect on the $MPI$ acquisition function within SEGOMOE.  }\label{fig:IGD:reg:MPI}
\end{figure}

\begin{figure}[H]
\begin{subfigure}{\textwidth}
\includegraphics[width=0.33\textwidth]{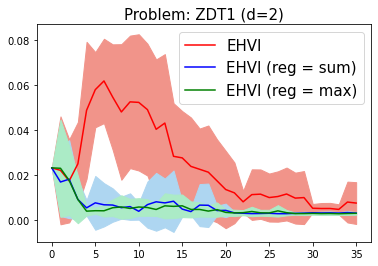}
\includegraphics[width=0.33\textwidth]{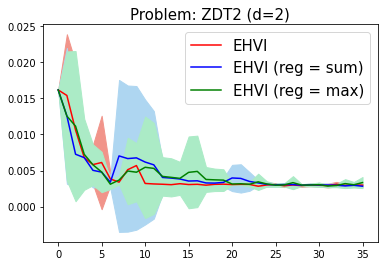}
\includegraphics[width=0.33\textwidth]{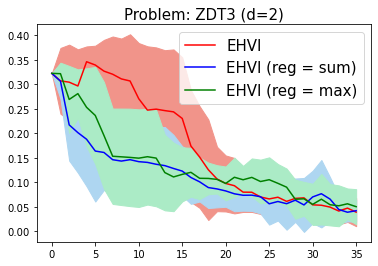}

\includegraphics[width=0.33\textwidth]{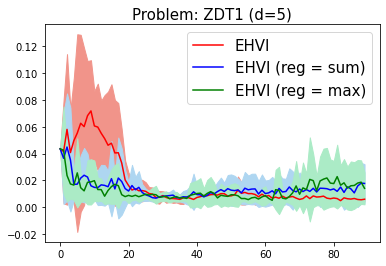}
\includegraphics[width=0.33\textwidth]{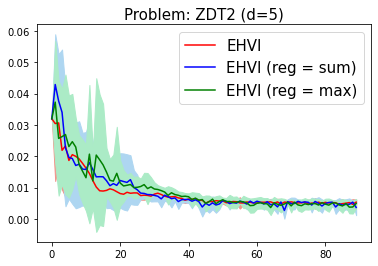}
\includegraphics[width=0.33\textwidth]{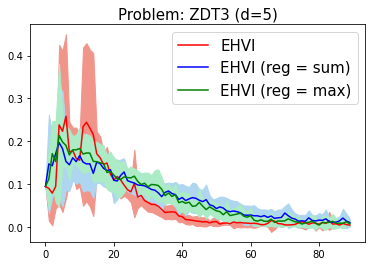}

\includegraphics[width=0.33\textwidth]{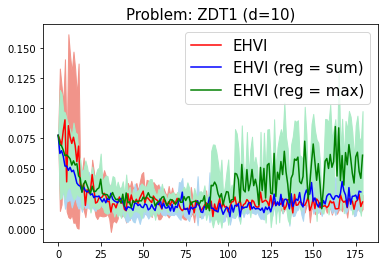}
\includegraphics[width=0.33\textwidth]{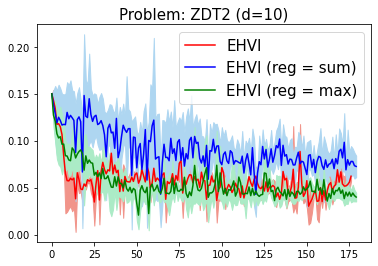}
\includegraphics[width=0.33\textwidth]{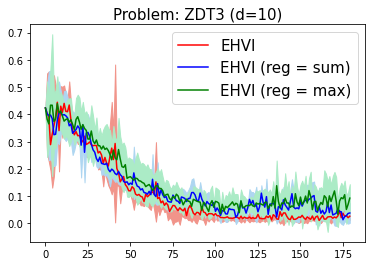}
\caption{Unconstrained problems (from left to right: ZDT1, ZDT2 and ZDT3). }\label{fig:uncons-IGD:reg:EHVI}
\end{subfigure}

\begin{subfigure}{\textwidth}
\includegraphics[width=0.33\textwidth]{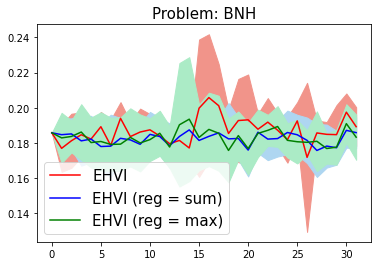}
\includegraphics[width=0.33\textwidth]{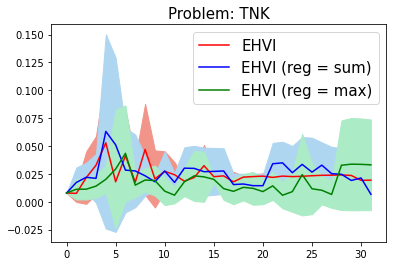}
\includegraphics[width=0.33\textwidth]{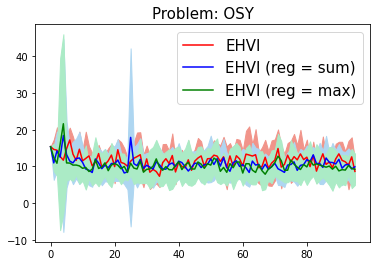}
\caption{Constrained problems (from left to right: BNH, TNK and OSY) }\label{fig:cons-IGD:reg:EHVI}
\end{subfigure}

\caption{Obtained convergence plots (i.e., ${IGD}^{+}$ values across iterations): regularization effect on the $EHVI$ acquisition function within SEGOMOE.  }\label{fig:IGD:reg:EHVI}
\end{figure}

The obtained convergence ${IGD}^{+}$ plots for the regularized infill criteria are given in Fig.~\ref{fig:IGD:reg:PI},~\ref{fig:IGD:reg:MPI} and~\ref{fig:IGD:reg:EHVI}, respectively, for the acquisition functions $PI$, $MPI$ and $EHVI$. 
The obtained results on our test cases using the regularized infill criteria are in average better all along the optimization steps and converge to smaller values of ${IGD}^{+}$ after the given budget of iterations. To measure both the quality of the obtained fronts after $20d$ iterations (our maximal evaluation budget), we compared the average of the ${IGD}^{+}$ scores across iterations in Table~\ref{tab:IGD-wb2s}. The gains are even more visible when the design space is high dimensional, thanks to the exploration improvement provided by the method. 
Nevertheless, even if $EHVI$ is the slowest infill criterion, it is the one giving the best results when regularized, especially with the transformation (reg = sum)  given by Eq.~(\ref{eq:sum}). 

As a conclusion from our numerical tests, in particular if the targeted problem is high dimensional ($d > 10$), we recommend to use the combination of $PI$ and the regularization based on the maximum of the mean prediction (i.e., reg = max).

\begin{table}[H]
\caption{\label{tab:IGD-wb2s} ${IGD}^{+}$  scores  (average of 10 runs) at the end of the optimization process, i.e., after $20d$ iterations. The best values are given in bold.}
  \begin{center}
   \resizebox{1.05\linewidth}{!}{
   \begin{tabular}{|c|ccc|ccc|ccc|ccc|}
   \hline
   \multirow{3}{*}{Acquisition func.} &\multicolumn{9}{c|}{Unconstrained problems} & \multicolumn{3}{c|}{Constrained problems} \\
   \cline{2-13}
   &\multicolumn{3}{c}{dimension 2}&\multicolumn{3}{c}{dimension 5} &\multicolumn{3}{c|}{dimension 10} & \multirow{2}{*}{BNH} & \multirow{2}{*}{TNK}&\multirow{2}{*}{OSY}  \\
   & ZDT1 & ZDT2 & ZDT3 & ZDT1 & ZDT2 & ZDT3 & ZDT1 & ZDT2 & ZDT3  &  &  &  \\
     \hline
     $PI$ & $2.50\, 10^{-2}$ & $3.7\, 10^{-3}$ & $1.79\, 10^{-1}$ &
     $2.23\, 10^{-2}$ & $8.5\, 10^{-3}$ & $\bf{6.19\, 10^{-2}}$ &
     $3.42\, 10^{-2}$ & $6.05\, 10^{-2}$ & $\bf{9.17\, 10^{-2}}$ & $\bf{1.82\, 10^{-1}}$ & $2.71\, 10^{-2}$ &$11.1$  \\ 
     $PI$ (reg = max) & $6.8\, 10^{-3}$ & $3.8\, 10^{-3}$ & $\bf{6.41\, 10^{-2}}$ &
     $9.0\, 10^{-3}$ & $\bf{7.3\, 10^{-3}}$ & $6.42\, 10^{-2}$ &
     $2.46\, 10^{-2}$ & $6.38\, 10^{-2}$ & $1.22\, 10^{-1}$ & $1.91\, 10^{-1}$ & $2.64\, 10^{-2}$ &$\bf{9.89 \,}$ \\ 
     $PI$ (reg = sum) &$\bf{5.9\, 10^{-3}}$ & $\bf{3.5\, 10^{-3}}$ & $2.45\, 10^{-1}$ &
     $\bf{8.2\, 10^{-3}}$ & $7.5\, 10^{-3}$ & $7.22\, 10^{-2}$ &
     $\bf{2.04\, 10^{-2}}$ & $\bf{4.68\, 10^{-2}}$ & $1.38\, 10^{-1}$ & $1.83\, 10^{-1}$ & $\bf{2.23\, 10^{-2}}$ &$11.2$ \\ 
     \hline
     $MPI$ & $1.27\, 10^{-2}$ & $3.8\, 10^{-3}$& $2.47\, 10^{-1}$ &
     $2.85\, 10^{-2}$ & $1.09\, 10^{-2}$ & $1.07\, 10^{-1}$ &
     $4.12\, 10^{-2}$&$6.09\, 10^{-2} $& $1.70\, 10^{-1}$&$\bf{1.83\, 10^{-1}}$ & $2.23\, 10^{-2}$ &$11.2$ \\ 
     $MPI$ (reg = max) &$1.06\, 10^{-2}$ & $5.2\, 10^{-3}$& $\bf{1.14\, 10^{-1}}$ &
     $1.26\, 10^{-2}$ & $\bf{8.1\, 10^{-3}}$ & $9.61\, 10^{-2}$ &
     $\bf{2.21\, 10^{-2}}$&$4.62\, 10^{-2} $& $1.82\, 10^{-1}$&$1.97\, 10^{-1}$ & $\bf{1.87\, 10^{-2}}$ &$\bf{1.11 \, 10^1}$\\
     $MPI$ (reg = sum) &$\bf{5.1\, 10^{-3}}$ & $\bf{3.7\, 10^{-3}}$& $1.64\, 10^{-1}$ &
     $\bf{7.3\, 10^{-3}}$ & $9.4\, 10^{-3}$ & $\bf{8.74\, 10^{-2}}$ &
     $2.34\, 10^{-2}$&$\bf{4.54\, 10^{-2}} $& $\bf{1.32\, 10^{-1}}$&$1.87\, 10^{-1}$ & $2.41\, 10^{-2}$ &$11.6$ \\
     \hline
     $EHVI$ & $2.47\, 10^{-2}$ & $4.0\, 10^{-3}$ & $1.70\, 10^{-1}$ &
     $1.76\, 10^{-2}$ & $\bf{8.4\, 10^{-3}}$ & $\bf{5.81\, 10^{-2}}$ &
     $2.55\, 10^{-2}$ & $5.78\, 10^{-2}$ & $\bf{1.13\, 10^{-1}}$& $\bf{1.82\, 10^{-1}}$ & $2.71\, 10^{-2}$ &$11.1$ \\ 
     $EHVI$ (reg = max) &$\bf{3.2\, 10^{-3}}$ & $\bf{3.5\, 10^{-3}}$ & $\bf{9.76\, 10^{-2}}$ &
     $1.39\, 10^{-2}$ & $9.0\, 10^{-2}$ & $7.13\, 10^{-2}$ &
     $\bf{2.39\, 10^{-2}}$ & $9.14\, 10^{-2}$ & $1.35\, 10^{-1}$&$\bf{1.82\, 10^{-1}}$ & $2.54\, 10^{-2}$ &$10.7$\\ 
     $EHVI$ (reg = sum) &$3.4\, 10^{-3}$ & $3.8\, 10^{-3}$ & $9.92\, 10^{-2}$ &
     $\bf{1.25\, 10^{-2}}$ & $9.4\, 10^{-2}$ & $7.05\, 10^{-2}$ &
     $3.73\, 10^{-2}$ & $\bf{5.49\, 10^{-2}}$ & $1.53\, 10^{-1}$&$1.83\, 10^{-1}$ & $\bf{1.75\, 10^{-2}}$ &$\bf{10.2}$ \\ 
    \hline
   \end{tabular}
   }
      \end{center}
   \end{table}

\section{Overall aircraft design bi-objective optimization}
\label{section 5}
The proposed approach is then applied to a MDO application from the FAST-OAD framework~\cite{David_2021}. FAST-OAD\footnote{\url{https://github.com/fast-aircraft-design/FAST-OAD}} is an open-source Python framework that provides a flexible way to build and solve the Overall Aircraft Design problems by assembling discipline models from various sources: FAST-OAD currently comes with some bundled, quick and simple, models dedicated to commercial aircraft. In this paper, FAST-OAD will resolve an MDA problem that mainly: (a) sizes the geometry of main aircraft components, (b) computes mass and centers of gravity of aircraft parts, (c)  estimates the aerodynamics and propulsion along the computed mission, and  (d) returns some quantities of interest related to the mission. These estimated quantities are used to define the two objectives and the constraints of our optimization problem. In this case, the fuel burn and the operating weight empty are considered as the two objectives.
The data from the CEntral Reference Aircraft System (``\texttt{CERAS}'') based on an Airbus A320 aircraft are used here. The problem to solve is a constrained optimization problem with 2 objective functions, 5 continuous design variables with associated bound constraints and 3 nonlinear inequality constraints.
The optimization problem is described in Tab.~\ref{tab:ceras}.

\begin{table}[H]
\vspace*{-0.2cm}
\centering
   \caption{Definition of the ``\texttt{CERAS}'' bi-objective optimization problem.}
\small
\resizebox{0.9\columnwidth}{!}{%
\small
\begin{tabular}{lclr}
 & Function/variable & Quantity  & Range\\
\hline
\hline
Minimize & Fuel mass  &  1 &\\
&  Operating Weight Empty & 1 \\
 \cline{2-4}
 & {\textbf{Total  objective  functions}} & {\textbf{2}} & \\
\hline
with respect to & \mbox{x position of Mean Aerodynamic Chord} &  1 & $\left[16., 18.\right]$ ($m$)\\
 & \mbox{Wing aspect ratio}  &1 & $\left[5., 11.\right]$ \\
 & \mbox{Horizontal tail aspect ratio}  & 1 & $\left[1.5, 6.\right]$ \\
  & \mbox{Wing taper aspect ratio} &  1 & $\left[0., 1.\right]$ \\
   & \mbox{Angle for swept wing} &  1 & $\left[20., 30.\right]$ ($^\circ$) \\
    \cline{2-4}
 & {\textbf{Total  design  variables}} & {\textbf{5}} & \\
 \cline{2-4}
  \hline
subject to & 0.05 \textless \ Static margin \textless \  0.1 & 2 \\
& Wing span \textless \  36  & 1\\
    \cline{2-4}
 & {\textbf{Total  constraints}} & {\textbf{3}} & \\
\hline
\end{tabular}
}
\label{tab:ceras}
\end{table}

\begin{figure}[H]
   \begin{subfigure}[b]{.5\linewidth}
      \centering
	\includegraphics[height=6cm]{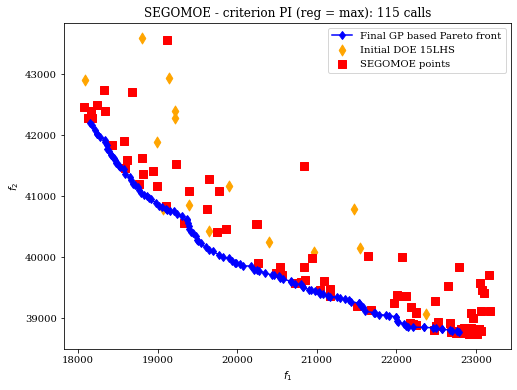}
     \caption{Approximate Pareto front obtained with SEGOMOE - PI (reg = max) criterion}
     \label{fig:pf_ceras_moo_PImax}
      \end{subfigure}
      \begin{subfigure}[b]{.5\linewidth}
      \centering 
\includegraphics[height=6cm]{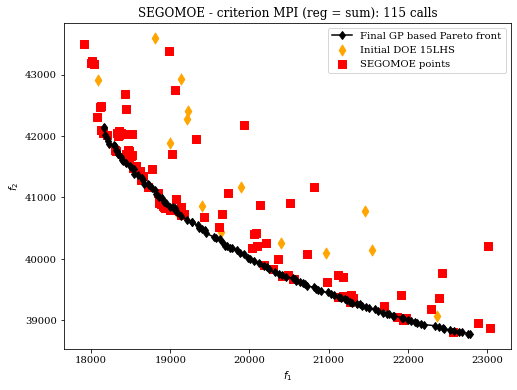}
 \caption{Approximate Pareto front obtained with SEGOMOE - $MPI$ (reg = sum) criterion }
\label{fig:pf_ceras_moo_MPIsum}
   \end{subfigure}
   \caption{``\texttt{CERAS}'' bi-objective optimization results with SEGOMOE considering an initial DoE of 15 points and 100 iterations. Two regularized acquisition criteria, $PI$ (reg = max) and $MPI$ (reg = max) are compared.}
   \label{fig:CeRAS_moo1}
\end{figure}

\begin{figure}[H]
   \begin{subfigure}[b]{.5\linewidth}
      \centering
	\includegraphics[height=6cm]{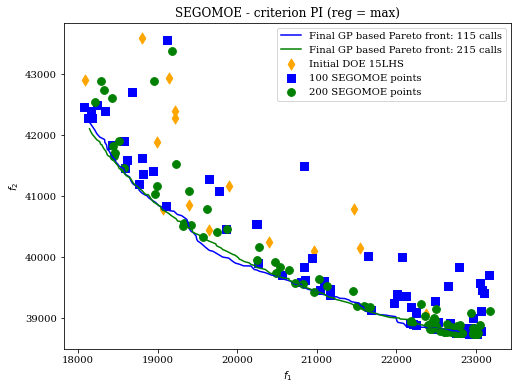}
     \caption{Approximate Pareto fronts obtained with SEGOMOE - $PI$ (reg = max) criterion: 115 and 215 calls }
     \label{fig:pf_ceras_moo_100_200iter}
      \end{subfigure}
      \begin{subfigure}[b]{.5\linewidth}
      \centering 
\includegraphics[height=6cm]{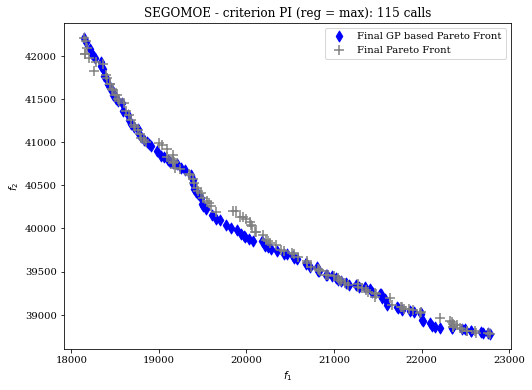}
 \caption{Approximate Pareto front (SEGOMOE - $PI$ (reg = max), 115 calls) and the recomputed one with FAST-OAD.}
\label{fig:pf_ceras_moo_exact}
   \end{subfigure}
   \caption{``\texttt{CERAS}'' bi-objective optimization results with SEGOMOE considering an initial DoE of 15 points and 100 or 200 iterations.}
   \label{fig:CeRAS_moo2}
\end{figure}

\begin{figure}[H]
   \begin{subfigure}[b]{.5\linewidth}
      \centering
	\includegraphics[height=6cm]{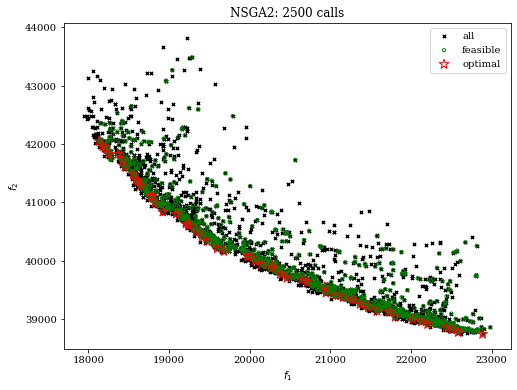}
     \caption{Admissible points, dominated points and Pareto front using NSGA-II. A total of 2500 points is represented.} \label{fig:pf_ceras_nsga2}
      \end{subfigure}
      \begin{subfigure}[b]{.5\linewidth}
      \centering 
\includegraphics[height=6cm]{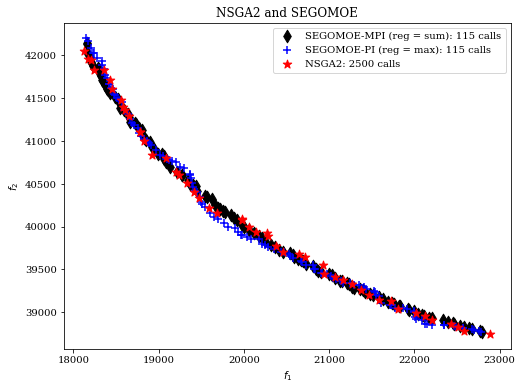}
 \caption{Comparison of Pareto fronts obtained with SEGOMOE (115 calls) and NSGA-II (2500 calls)}
\label{fig:pf_ceras_moo_nsga2_exact}
   \end{subfigure}
   \caption{``\texttt{CERAS}'' bi-objective optimization results with SEGOMOE and NSGA-II.}
   \label{fig:CeRAS_moo_nsga2}
\end{figure}

The proposed method SEGOMOE for Multi-Objective Bayesian Optimization, is applied on the  ``\texttt{CERAS}'' problem with an initial DoE of 15 points and 100 iterations are performed. For these preliminary results, two different  acquisition functions with regularization, $PI$ (reg = max) and $MPI$ (reg = sum) are compared and a population size of 120 points is chosen for the approximate Pareto front. The resulting Pareto fronts are presented in Fig.~\ref{fig:pf_ceras_moo_PImax} and Fig.~\ref{fig:pf_ceras_moo_MPIsum} with the initial DoE and the 100 added BO points. The approximate Pareto front is composed of 120 points obtained with the GP surrogate models (black or blue points depending on the used criterion). 
If more iterations are performed, the Pareto fronts are quite similar: Fig.~\ref{fig:pf_ceras_moo_100_200iter} compares the optimal fronts for  PI (reg=max) criterion considering 100 or 200 iterations. 
In order to validate the SEGOMOE Pareto front obtained, we compare the 120 GP values (approximate Pareto front with blue points) to the real ones (grey points)  computed directly with FAST-OAD in Fig~\ref{fig:pf_ceras_moo_exact}. All of the 120 points respect the constraints and the two Pareto fronts are closed to each other which confirms the accuracy of the GP models.
The last comparison is done by applying directly the NSGA-II algorithm on the FAST-OAD framework without using any surrogate model on Fig.~\ref{fig:CeRAS_moo_nsga2}. To run the algorithm, 50 generations are used with a population size equals to 50. Fig.~\ref{fig:pf_ceras_nsga2} shows the 2500 obtained points where the green ones represent the feasible configurations and the red ones the non dominated solutions. The SEGOMOE (approximate with $PI$ (reg = max) or $MPI$ (reg = sum)) and the NSGA-II Pareto fronts are finally compared on  Fig.~\ref{fig:pf_ceras_moo_nsga2_exact}. In terms of number of function evaluations, the SEGOMOE front is obtained with 115 evaluations of FAST-OAD compared to 2500 evaluations with NSGA-II.


Concerning the computational time of this test case, Table~\ref{tab:CPUtime} compares the CPU time required to obtain the different final Pareto fronts of Fig~\ref{fig:CeRAS_moo_nsga2}. The SEGOMOE algorithm with both selecting criteria ($PI$ (reg = max)  or  $MPI$ (reg = sum))  allows to reduce drastically the cost: less than 5 hours instead of 85 hours with NSGA-II. The CPU time for the $PI$ criterion (reg = max) is divided by a factor of 1.3 compared to the $MPI$ criterion (reg = sum).
\begin{table}[]
    \centering
    \begin{tabular}{c|c|c}
         Approaches& Number of calls & Time  \\
         \hline
         NSGA-II& 2500 & 85h \\
         SEGOMOE - $PI$ (reg = max) & 115 & 3h10 \\
         SEGOMOE - $MPI$ (reg = sum) & 115 & 4h04 \\
    \end{tabular}
    \caption{CPU time to compare the different approaches.}
    \label{tab:CPUtime}
\end{table}

\begin{figure}[H]
   \begin{subfigure}[b]{.5\linewidth}
      \centering
	\includegraphics[height=6cm]{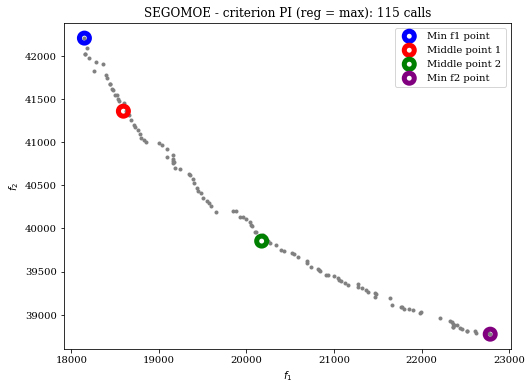}
     \caption{4 selected points on the Pareto front.}
     \label{fig:4_ceras_points}
      \end{subfigure}
      \begin{subfigure}[b]{.5\linewidth}
      \centering 
\includegraphics[height=5.5cm,width=8cm]{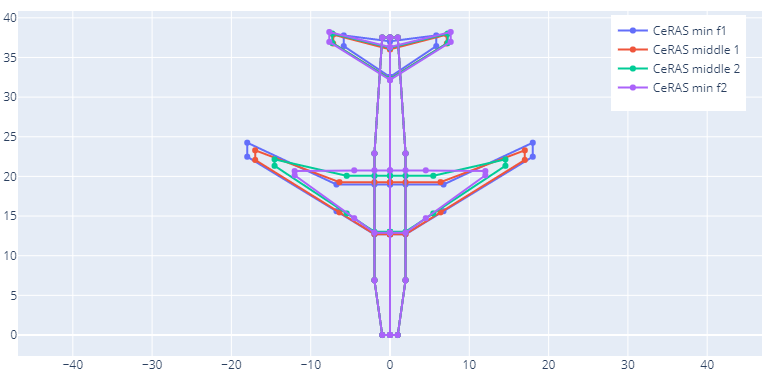}
\vspace{.25cm}
 \caption{`\texttt{CERAS}'' geometry for four specific points}
\label{fig:4_ceras_geometry}
   \end{subfigure}
   \caption{``\texttt{CERAS}'' geometry configurations for four selected points on the Pareto front.}
   \label{fig:CeRAS_geometry}
\end{figure}

\begin{figure}[H]

      \centering
	\includegraphics[width= \linewidth,height=7cm]{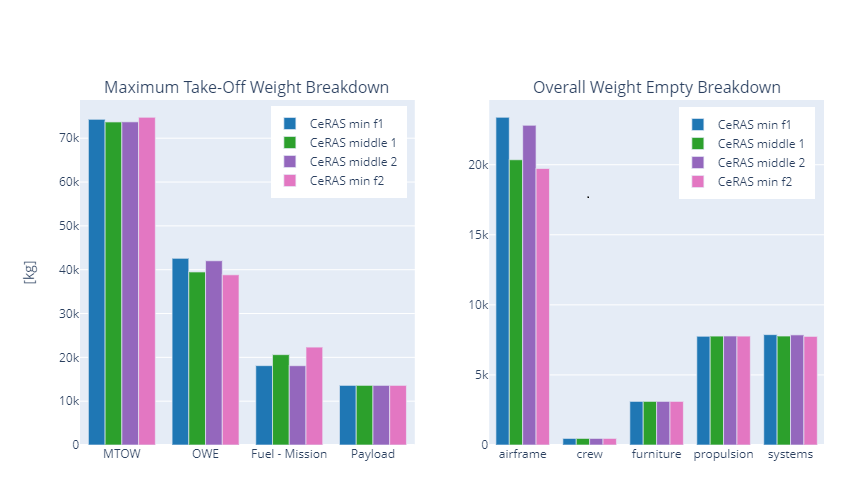}
     \caption{Mass breakdown: Maximum TakeOff Weight and Overall Weight Empty for the four selected ``\texttt{CERAS}'' different configurations.}
     \label{fig:4_ceras_points_mass}
      \end{figure}

For illustration, four points have been selected on the Pareto front and their associated ``\texttt{CERAS}'' geometries are represented in Fig.~\ref{fig:CeRAS_geometry}: the first point is given by the minimum of fuel burn ($\min f_1$),  two points are located  along the front, and the forth  one is associated to the minimum of Operating Weight Empty (OWE) ($\min f_2$) as shown in  Fig.~\ref{fig:4_ceras_points}. The differences between the four resulting configurations are illustrated in  Fig.~\ref{fig:4_ceras_geometry}. The minimum of fuel burn ($\min f_1$) configuration exhibits the highest aspect ratio which is beneficial for (induced) drag reduction but leads to a higher structural weight. Fig.~\ref{fig:4_ceras_points} shows that this configuration has an increased OWE (including wing structural weight) but also a reduced fuel burn. On the opposite, the minimum OWE ($\min f_2$) has a more compact wing geometry close to a delta wing. Such a geometry has a lower structural weight associated with a lower aerodynamic performance. The impact on OWE and fuel burn is confirmed by Fig.~\ref{fig:4_ceras_points_mass}:  although the simplified aerodynamic and wing mass models are at the limit of their validity for such a geometry, their ability to capture the expected tendencies is satisfactory for this study.

      %
\section{Conclusion and perspectives}
\label{section 6}
In this paper, we extended the BO optimizer SEGOMOE to solve constrained multi-objective problems. The use of different acquisition functions within SEGOMOE was also investigated. A class of regularized infill criteria for the acquisition functions  have been proposed in order to ease the optimization process in a multi-objective context.
The merit of our approach was first shown on known analytical multi-objective problems with and without constraints. Then, the proposed method was used to solve a bi-objective application related to conceptual aircraft design based on the Airbus A320 aircraft data. Including categorical design variables will allow to solve more realistic conceptual aircraft design problems~\cite{dlr2022}. In this context, inspired by the recent software developments~\cite{SciTech_cat,Aerobest_cat}, an extension of SEGOMOE to solve mixed-categorical multi-objective optimization problems shall be investigated in a near future. 


\section{ Appendix}

In this appendix, we describe mathematically the benchmark problems we used to test our implementations. 

\subsection{ Unconstrained problems}\label{appendix1}
The unconstrained test cases are the three first ZDT functions~\cite{ZDT}. All of them can take at least $d=2$ variables. Their domain is then $\Omega = [0,1]^d$.  Their respective expressions are:

$ZDT1 : \x \in \mathbb{R}^d \mapsto [x_1 , 1 - \sqrt{\frac{x_1}{\frac{\sum_{i = 2}^d x_i}{d-1}}}]$

$ZDT2 : \x \in \mathbb{R}^d  \mapsto [x_1 , 1 - \left(\frac{x_1}{\frac{\sum_{i = 2}^d x_i}{d-1}}\right)^2]$

$ZDT3 : \x \in \mathbb{R}^d  \mapsto [x_1 , 1 - \sqrt{\frac{x_1}{1 + \frac{\sum_{i = 2}^d x_i}{d-1}}} - \frac{x_1}{1 + \frac{\sum_{i = 2}^d x_i}{d-1}}\sin(10\pi x_1)].$

\begin{figure}[H]
\centering
\includegraphics[width=0.3\textwidth]{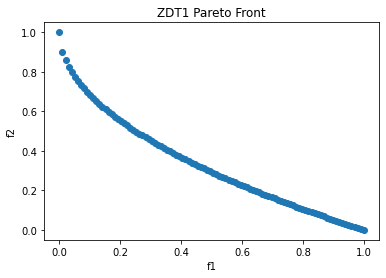}%
\includegraphics[width=0.3\textwidth]{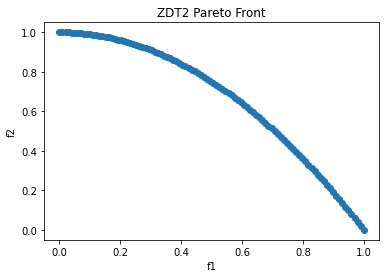}%
\includegraphics[width=0.3\textwidth]{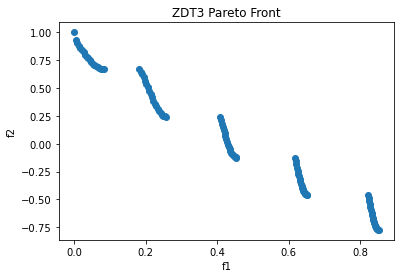}%
\caption{The optimal Pareto fronts of the tested unconstrained multi-objective problems: ZDT1, ZDT2 and ZDT3.} \label{fig:analytZDTfront}
\end{figure}

These functions are commonly used in the literature thanks to the variety of Pareto fronts they provide. ZDT1 has a convex Pareto front, ZDT2 has a concave one and ZDT3 a discontinuous front as shown in Fig.~\ref{fig:analytZDTfront}. As the coordinate $x_1$ has opposite effects in both objectives, the Pareto optimal points are obtained by minimizing $g$, such that the Pareto set is $\{ x \in \Omega \; ; \; \forall i > 1, x_i = 0 \}$. 

\subsection{ Constrained problems}\label{appendix2}

\begin{figure}[H]
\centering
\includegraphics[width=0.3\textwidth]{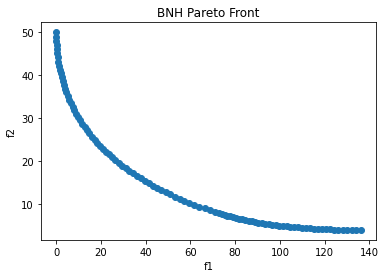}%
\includegraphics[width=0.3\textwidth]{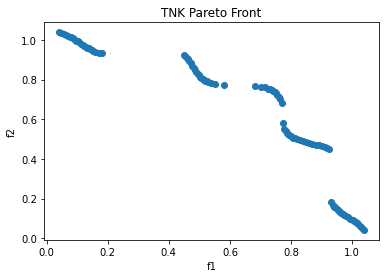}%
\includegraphics[width=0.3\textwidth]{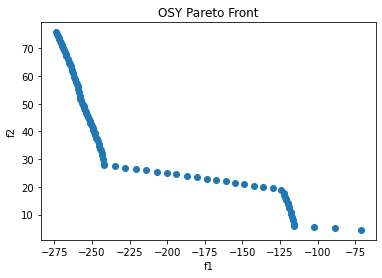}%
\caption{The optimal Pareto fronts of the tested constrained multi-objective problems: BNH, TNK and OSY.} 
\label{fig:analytCONSTpf}
\end{figure}

\begin{itemize}
    \item The first one is the Binh and Korn problem~\cite{binh1997mobes}.
\begin{equation}
\f:
\left.
  \begin{array}{rcl}
  \mathbb{R}^2 & \longrightarrow &\mathbb{R}^2 \\
    \boldx & \longmapsto & [4x_1^2 + 4x_2^2 \;, \; (x_1-5)^2 + (x_2-5)^2]\\
  \end{array}
\right.\\
\end{equation}
\begin{equation}
\mbox{ such that }  
\left \{
\begin{array}{rcl}
 g_1(\boldx) &=& -(x_1-5)^2 - x_2^2 +25 \geq 0 \\
g_2(\boldx) &=& (x_1-8)^2 + (x_2+3)^2 -7.7 \geq 0 \\
\end{array}
\right.
\end{equation}
The design space is given by $\Omega=   [0,5]\times[0,3].$
Its Pareto set is the following union $\{ \boldx \; ;\; x_1 = x_2 \;\mbox{ and }\; x_1 \in [0,3]\} \bigcup \{x \; ; \; x_1 \in [3,5] \;\mbox{ and }\; x_2 = 3 \}$.
\item The second problem is the Tanaka problem~\cite{zapotecas2018review} with 2 variables and 2 constraints:
\begin{equation}
\f:
\left.
  \begin{array}{rcl}
  \mathbb{R}^2 & \longrightarrow &\mathbb{R}^2 \\
    \boldx & \longmapsto & [x_1, x_2]\\
  \end{array}
\right.\\
\end{equation}
\begin{equation}
\begin{array}{rcl}
\mbox{ such that }  \;
\left\{
    \begin{array}{lll}
     g_1(\boldx) &=& x_1^2 + x_2^2 - 1 -
        0.1\cos \left(16\arctan \frac{x_1}{x_2}\right) \geq 0 \\ 
        g_2(\boldx) &=& - (x_1-0.5)^2 - (x_2-0.5)^2 + 0.5 \geq 0\\  
    \end{array}
\right.

\end{array}
\end{equation}
The design space is given by $\Omega=   [0,\pi]\times[0,\pi]$. If there is no constraint, the solution would be $\boldx = (0,0)$. As this point is not reachable, the solution lies on the boundary of the first constraint for the points that also satisfy the second constraint, making of the Pareto optima a discontinuous set as shown on Fig.~\ref{fig:analytCONSTpf}.
\item The third constrained problem is from Osyczka and Kundu paper~\cite{osyczka1995new}: 
\begin{equation}
\f:
\left.
  \begin{array}{rcl}
    \mathbb{R}^6 & \longrightarrow &\mathbb{R}^2 \\
    \boldx & \longmapsto & [-(25(x_1-2)^2+(x_2-2)^2 + (x_3-1)^2+(x_4-4)^2  + (x_5-1)^2) \;, x_1^2 + x_2^2 + x_3^2 + x_4^2 + x_5^2 + x_6^2]\\
  \end{array}
\right.\\
\end{equation}
\begin{equation}
\mbox{ such that }  \;
\left\{
\begin{array}{rcl}
g_1(\boldx) &=& x_1 + x_2 - 2 \geq 0 \\
g_2(\boldx) &=& 6 - x_1 - x_2 \geq 0 \\
 g_3(\boldx) &=& 2 - x_2 + x_1 \geq 0 \\
 g_4(\boldx) &=& 2 - x_1 + 3x_2 \geq 0 \\
 g_5(\boldx) &=& 4 - (x_3-3)^2 - x_4 \geq 0 \\
 g_6(\boldx) &=& (x_5-3)^2 + x_6 - 4 \geq 0 \\
\end{array}
\right.
\end{equation}

The design space is given by $\Omega=[0,10]\times[0,10]\times[0,5]\times[0,6]\times[0,5]\times[0,10].$
This problem has the following explicit optimal Pareto front set
\begin{center}
   \begin{tabular}{cccccc}
    $x_1$ & $x_2$ &$ x_3$ & $x_4$ & $x_5$ & $x_6$   \\
     \hline
     5 & 1 & [1,5] & 0 & 5 & 0\\ 
     5 & 1 & [1,5] & 0 & 1 & 0\\ 
     $[4.065, 5]$ & $(x_1 - 2)/3$ & 1 & 0 & 1 & 0 \\
     0 & 2 & $[1,1.3732]$ & 0 & 1 & 0 \\ 
     $[0, 1]$ & $2 - x_1$ & 1 & 0 & 1 & 0\\ 
   \end{tabular}
 \end{center}

\end{itemize}
The problems NH, TNK and OSY have  different Pareto front structures are shown in Fig.~\ref{fig:analytCONSTpf}. 

\section*{Acknowledgments}
This work is part of the activities of ONERA - ISAE - ENAC joint research group. 
The research presented in this paper has been performed in the framework of the AGILE 4.0 project (Towards
Cyber-physical Collaborative Aircraft Development) and has received funding from the European Union Horizon 2020
Programme under grant agreement n\textsuperscript{o}~815122.
The authors would like to thank Christophe David and Sébastien Defoort for their support with the framework FAST-OAD concerning the application test case. 

\bibliography{main}

\begin{thebibliography}{48}
\newcommand{\enquote}[1]{``#1''}
\providecommand{\natexlab}[1]{#1}
\providecommand{\url}[1]{\texttt{#1}}
\providecommand{\urlprefix}{URL }
\expandafter\ifx\csname urlstyle\endcsname\relax
  \providecommand{\doi}[1]{\discretionary{}{}{}https://doi.org/#1}\else
  \providecommand{\doi}[1]{\discretionary{}{}{}\urlstyle{rm}\url{https://doi.org/#1}}\fi

\bibitem[{Raymer(2018)}]{RaymerAircraftDesignConceptual2018}
Raymer, D.~P., \emph{Aircraft Design: A Conceptual Approach, Sixth Edition}, {Virginia: American Institute of Aeronautics \& Astronautics}, 2018.

\bibitem[{Priem et~al.(2020{\natexlab{a}})Priem, Gagnon, Chittick, Dufresne, Diouane, and Bartoli}]{BRAC_AIAA:20}
Priem, R., Gagnon, H., Chittick, I., Dufresne, S., Diouane, Y., and Bartoli, N., \enquote{An efficient application of Bayesian optimization to an industrial MDO framework for aircraft design.} \emph{AIAA AVIATION 2020 FORUM}, 2020{\natexlab{a}}, p. 3152.

\bibitem[{Mo{\v c}kus(1975)}]{MockusBayesianmethodsseeking1975}
Mo{\v c}kus, J., \enquote{On {{Bayesian}} Methods for Seeking the Extremum,} \emph{Optimization {{Techniques IFIP Technical Conference}}}, 1975, pp. 400--404.

\bibitem[{Jones et~al.(1998)Jones, Schonlau, and Welch}]{JonesEfficientglobaloptimization1998}
Jones, D.~R., Schonlau, M., and Welch, W.~J., \enquote{Efficient Global Optimization of Expensive Black-Box Functions,} \emph{Journal of Global optimization}, Vol.~13, 1998, pp. 455--492.

\bibitem[{Bartoli et~al.(2019{\natexlab{a}})Bartoli, Lefebvre, Dubreuil, Olivanti, Priem, Bons, Martins, and Morlier}]{wb2s}
Bartoli, N., Lefebvre, T., Dubreuil, S., Olivanti, R., Priem, R., Bons, N., Martins, J.~R., and Morlier, J., \enquote{Adaptive modeling strategy for constrained global optimization with application to aerodynamic wing design,} \emph{Aerospace Science and technology}, Vol.~90, 2019{\natexlab{a}}, pp. 85--102.

\bibitem[{Bettebghor et~al.(2011)Bettebghor, Bartoli, Grihon, Morlier, and Samuelides}]{bettebghor2011surrogate}
Bettebghor, D., Bartoli, N., Grihon, S., Morlier, J., and Samuelides, M., \enquote{Surrogate modeling approximation using a mixture of experts based on EM joint estimation,} \emph{Structural and multidisciplinary optimization}, Vol.~43, No.~2, 2011, pp. 243--259.

\bibitem[{Priem et~al.(2019)Priem, Bartoli, and Diouane}]{priem2019use}
Priem, R., Bartoli, N., and Diouane, Y., \enquote{On the Use of Upper Trust Bounds in Constrained Bayesian Optimization Infill Criteria,} \emph{AIAA Aviation 2019 Forum}, 2019, p. 2986.

\bibitem[{Priem et~al.(2020{\natexlab{b}})Priem, Bartoli, Diouane, and Sgueglia}]{priem2020upper}
Priem, R., Bartoli, N., Diouane, Y., and Sgueglia, A., \enquote{Upper trust bound feasibility criterion for mixed constrained Bayesian optimization with application to aircraft design,} \emph{Aerospace Science and Technology}, 2020{\natexlab{b}}, p. 105980.

\bibitem[{Feliot et~al.(2017)Feliot, Bect, and Vazquez}]{FeliotBayesianapproachconstrained2017}
Feliot, P., Bect, J., and Vazquez, E., \enquote{A {{Bayesian}} Approach to Constrained Single-and Multi-Objective Optimization,} \emph{Journal of Global Optimization}, Vol.~67, 2017, pp. 97--133.

\bibitem[{Knowles(2006)}]{Knowles_2006}
Knowles, J., \enquote{ParEGO: a hybrid algorithm with on-line landscape approximation for expensive multiobjective optimization problems,} \emph{IEEE Transactions on Evolutionary Computation}, Vol.~10, No.~1, 2006, pp. 50--66.
\newblock \doi{10.1109/TEVC.2005.851274}.

\bibitem[{Zitzler et~al.(2003)Zitzler, Thiele, Laumanns, Fonseca, and da~Fonseca}]{Zitzler_etal_2003}
Zitzler, E., Thiele, L., Laumanns, M., Fonseca, C., and da~Fonseca, V., \enquote{Performance assessment of multiobjective optimizers: an analysis and review,} \emph{IEEE Transactions on Evolutionary Computation}, Vol.~7, No.~2, 2003, pp. 117--132.
\newblock \doi{10.1109/TEVC.2003.810758}.

\bibitem[{Wagner et~al.(2010)Wagner, Emmerich, Deutz, and Ponweiser}]{Wagner_etal_2010}
Wagner, T., Emmerich, M., Deutz, A., and Ponweiser, W., \enquote{On Expected-Improvement Criteria for Model-based Multi-objective Optimization,} \emph{Parallel Problem Solving from Nature, PPSN XI}, edited by R.~Schaefer, C.~Cotta, J.~Ko{\l}odziej, and G.~Rudolph, Springer Berlin Heidelberg, Berlin, Heidelberg, 2010, pp. 718--727.

\bibitem[{Ponweiser et~al.(2008)Ponweiser, Wagner, Biermann, and Vincze}]{Ponweiser_etal_2008}
Ponweiser, W., Wagner, T., Biermann, D., and Vincze, M., \enquote{Multiobjective Optimization on a Limited Budget of Evaluations Using Model-Assisted S-Metric Selection,} \emph{Parallel Problem Solving from Nature -- PPSN X}, edited by G.~Rudolph, T.~Jansen, N.~Beume, S.~Lucas, and C.~Poloni, Springer Berlin Heidelberg, Berlin, Heidelberg, 2008, pp. 784--794.

\bibitem[{Emmerich et~al.(2006)Emmerich, Giannakoglou, and Naujoks}]{Emmerich_etal_2006}
Emmerich, M., Giannakoglou, K., and Naujoks, B., \enquote{Single- and multiobjective evolutionary optimization assisted by Gaussian random field metamodels,} \emph{IEEE Transactions on Evolutionary Computation}, Vol.~10, No.~4, 2006, pp. 421--439.
\newblock \doi{10.1109/TEVC.2005.859463}.

\bibitem[{Rahat et~al.(2017)Rahat, Everson, and Fieldsend}]{Rahat_etal_2017}
Rahat, A. A.~M., Everson, R.~M., and Fieldsend, J.~E., \enquote{Alternative Infill Strategies for Expensive Multi-Objective Optimisation,} \emph{Proceedings of the Genetic and Evolutionary Computation Conference}, Association for Computing Machinery, New York, NY, USA, 2017, p. 873–880.

\bibitem[{Zuluaga et~al.(2013)Zuluaga, Sergent, Krause, and Püschel}]{Zuluaga_etal_2013}
Zuluaga, M., Sergent, G., Krause, A., and Püschel, M., \enquote{Active Learning for Multi-Objective Optimization,} \emph{Proceedings of the 30th International Conference on Machine Learning}, Proceedings of Machine Learning Research, Vol.~28, edited by S.~Dasgupta and D.~McAllester, PMLR, Atlanta, Georgia, USA, 2013, pp. 462--470.

\bibitem[{Zuluaga et~al.(2016)Zuluaga, Krause, and P{{{\"u}}}schel}]{Zuluaga_etal_2016}
Zuluaga, M., Krause, A., and P{{{\"u}}}schel, M., \enquote{e-PAL: An Active Learning Approach to the Multi-Objective Optimization Problem,} \emph{Journal of Machine Learning Research}, Vol.~17, No. 104, 2016, pp. 1--32.

\bibitem[{Jones(2001)}]{Jones_2001}
Jones, D.~R., \enquote{A Taxonomy of Global Optimization Methods Based on Response Surfaces,} \emph{Journal of Global Optimization}, Vol.~21, 2001, pp. 345--383.

\bibitem[{Picheny(2014)}]{Picheny_2014}
Picheny, V., \enquote{{A Stepwise uncertainty reduction approach to constrained global optimization},} \emph{Proceedings of the Seventeenth International Conference on Artificial Intelligence and Statistics}, Proceedings of Machine Learning Research, Vol.~33, edited by S.~Kaski and J.~Corander, PMLR, Reykjavik, Iceland, 2014, pp. 787--795.

\bibitem[{Picheny(2015)}]{Picheny_2015}
Picheny, V., \enquote{Multiobjective optimization using Gaussian process emulators via stepwise uncertainty reduction,} \emph{Statistics and Computing}, Vol.~25, No.~6, 2015, pp. 1265--1280.

\bibitem[{Watson and Barnes(1995)}]{watson1995infill}
Watson, A.~G., and Barnes, R.~J., \enquote{Infill sampling criteria to locate extremes,} \emph{Mathematical Geology}, Vol.~27, No.~5, 1995, pp. 589--608.

\bibitem[{Rasmussen and Williams(2006)}]{RasmussenGaussianprocessesmachine2006}
Rasmussen, C.~E., and Williams, C. K.~I., \emph{Gaussian Processes for Machine Learning}, {The MIT Press}, 2006.

\bibitem[{Krige(1951)}]{Krigestatisticalapproachbasic1951}
Krige, D.~G., \enquote{A Statistical Approach to Some Basic Mine Valuation Problems on the {{Witwatersrand}},} \emph{Journal of the Southern African Institute of Mining and Metallurgy}, Vol.~52, 1951, pp. 119--139.

\bibitem[{Frazier(2018)}]{frazier2018tutorial}
Frazier, P.~I., \enquote{A tutorial on Bayesian optimization,} \emph{aarXiv:1807.02811}, 2018.

\bibitem[{Bartoli et~al.(2019{\natexlab{b}})Bartoli, Lefebvre, Dubreuil, Olivanti, Priem, Bons, Martins, and Morlier}]{Bartoliadaptivemodeling2019}
Bartoli, N., Lefebvre, T., Dubreuil, S., Olivanti, R., Priem, R., Bons, N., Martins, J., and Morlier, J., \enquote{Adaptive modeling strategy for constrained global optimization with application to aerodynamic wing design,} \emph{Aerospace Science and technology}, Vol.~90, 2019{\natexlab{b}}, pp. 85--102.

\bibitem[{Wang and Jegelka(2017)}]{WangMaxvalueentropysearch2017}
Wang, Z., and Jegelka, S., \enquote{Max-Value Entropy Search for Efficient {{Bayesian}} Optimization,} \emph{34th {{International Conference}} on {{Machine Learning}}}, Vol.~70, 2017, pp. 3627--3635.

\bibitem[{Lam et~al.(2015)Lam, Allaire, and Willcox}]{lam2015multifidelity}
Lam, R., Allaire, D., and Willcox, K., \enquote{Multifidelity optimization using statistical surrogate modeling for non-hierarchical information sources,} \emph{56th AIAA/ASCE/AHS/ASC Structures, Structural Dynamics, and Materials Conference}, 2015, p. 143.

\bibitem[{Sasena et~al.(2002)Sasena, Papalambros, and Goovaerts}]{SasenaExplorationmetamodelingsampling2002}
Sasena, M.~J., Papalambros, P., and Goovaerts, P., \enquote{Exploration of Metamodeling Sampling Criteria for Constrained Global Optimization,} \emph{Engineering optimization}, Vol.~34, 2002, pp. 263--278.

\bibitem[{Bouhlel et~al.(2016)Bouhlel, Bartoli, Otsmane, and Morlier}]{BouhlelImprovingkrigingsurrogates2016}
Bouhlel, M.~A., Bartoli, N., Otsmane, A., and Morlier, J., \enquote{Improving Kriging Surrogates of High-Dimensional Design Models by {{Partial Least Squares}} Dimension Reduction,} \emph{Structural and Multidisciplinary Optimization}, Vol.~53, 2016, pp. 935--952.

\bibitem[{Bouhlel et~al.(2019)Bouhlel, Hwang, Bartoli, Lafage, Morlier, and Martins}]{smt}
Bouhlel, M.~A., Hwang, J.~T., Bartoli, N., Lafage, R., Morlier, J., and Martins, J.~R., \enquote{A Python surrogate modeling framework with derivatives,} \emph{Advances in Engineering Software}, Vol. 135, 2019, p. 102662.

\bibitem[{Powell(1994)}]{cobyla}
Powell, M.~J., \enquote{A direct search optimization method that models the objective and constraint functions by linear interpolation,} \emph{Advances in optimization and numerical analysis}, Springer, 1994, pp. 51--67.

\bibitem[{Kraft et~al.(1988)}]{slsqp}
Kraft, D., et~al., \emph{A software package for sequential quadratic programming}, DFVLR Obersfaffeuhofen, Germany, 1988.

\bibitem[{Gill et~al.(2005)Gill, Murray, and Saunders}]{snopt}
Gill, P.~E., Murray, W., and Saunders, M.~A., \enquote{SNOPT: An SQP algorithm for large-scale constrained optimization,} \emph{SIAM review}, Vol.~47, No.~1, 2005, pp. 99--131.

\bibitem[{Bartoli et~al.(2016)Bartoli, Bouhlel, Kurek, Lafage, Lefebvre, Morlier, Priem, Stilz, and Regis}]{SEGOMOE_ISSMO:16}
Bartoli, N., Bouhlel, M.-A., Kurek, I., Lafage, R., Lefebvre, T., Morlier, J., Priem, R., Stilz, V., and Regis, R., \enquote{Improvement of efficient global optimization with application to aircraft wing design,} \emph{17th AIAA/ISSMO Multidisciplinary Analysis and Optimization Conference}, Washington D.C., USA, 2016, p. 4001.
\newblock \doi{10.2514/6.2016-4001}.

\bibitem[{Bartoli et~al.(2017)Bartoli, Lefebvre, Dubreuil, Olivanti, Bons, Martins, Bouhlel, and Morlier}]{SEGOMOE_AIAA:17}
Bartoli, N., Lefebvre, T., Dubreuil, S., Olivanti, R., Bons, N., Martins, J., Bouhlel, M.-A., and Morlier, J., \enquote{An adaptive optimization strategy based on mixture of experts for wing aerodynamic design optimization,} \emph{18th AIAA/ISSMO Multidisciplinary Analysis and Optimization Conference}, 2017, p. 4433.
\newblock \doi{10.2514/6.2017-4433}.

\bibitem[{Savelyev et~al.(2018)Savelyev, Anisomov, Bartoli, Lefebvre, Dubreuil, Panzeri, and D'Ippolito}]{AGILE_Nacelle_OneraTsagi:18}
Savelyev, A., Anisomov, K., Bartoli, N., Lefebvre, T., Dubreuil, S., Panzeri, M., and D'Ippolito, R., \enquote{Robust design optimization applied to turbofan nacelle,} \emph{17th TsAGI-ONERA Seminar}, Moscow, Russia, 2018.

\bibitem[{Deb et~al.(2002)Deb, Pratap, Agarwal, and Meyarivan}]{nsga2}
Deb, K., Pratap, A., Agarwal, S., and Meyarivan, T., \enquote{A fast and elitist multiobjective genetic algorithm: NSGA-II,} \emph{IEEE transactions on evolutionary computation}, Vol.~6, No.~2, 2002, pp. 182--197.

\bibitem[{Zitzler et~al.(2000)Zitzler, Deb, and Thiele}]{ZDT}
Zitzler, E., Deb, K., and Thiele, L., \enquote{Comparison of multiobjective evolutionary algorithms: Empirical results,} \emph{Evolutionary computation}, Vol.~8, No.~2, 2000, pp. 173--195.

\bibitem[{Binh and Korn(1997)}]{binh1997mobes}
Binh, T.~T., and Korn, U., \enquote{MOBES: A multiobjective evolution strategy for constrained optimization problems,} \emph{The third international conference on genetic algorithms (Mendel 97)}, Vol.~25, Citeseer, 1997, p.~27.

\bibitem[{Zapotecas-Mart{\'\i}nez et~al.(2018)Zapotecas-Mart{\'\i}nez, Coello, Aguirre, and Tanaka}]{zapotecas2018review}
Zapotecas-Mart{\'\i}nez, S., Coello, C. A.~C., Aguirre, H.~E., and Tanaka, K., \enquote{A review of features and limitations of existing scalable multiobjective test suites,} \emph{IEEE Transactions on Evolutionary Computation}, Vol.~23, No.~1, 2018, pp. 130--142.

\bibitem[{Osyczka and Kundu(1995)}]{osyczka1995new}
Osyczka, A., and Kundu, S., \enquote{A new method to solve generalized multicriteria optimization problems using the simple genetic algorithm,} \emph{Structural optimization}, Vol.~10, No.~2, 1995, pp. 94--99.

\bibitem[{Jin et~al.(2003)Jin, Chen, and Sudjianto}]{lhs}
Jin, R., Chen, W., and Sudjianto, A., \enquote{An efficient algorithm for constructing optimal design of computer experiments,} \emph{International Design Engineering Technical Conferences and Computers and Information in Engineering Conference}, Vol. 37009, 2003, pp. 545--554.

\bibitem[{{Blank} and {Deb}(2020)}]{pymoo}
{Blank}, J., and {Deb}, K., \enquote{Pymoo: Multi-Objective Optimization in Python,} \emph{IEEE Access}, Vol.~8, 2020, pp. 89497--89509.

\bibitem[{Ishibuchi et~al.(2015)Ishibuchi, Masuda, Tanigaki, and Nojima}]{ishibuchi2015modified}
Ishibuchi, H., Masuda, H., Tanigaki, Y., and Nojima, Y., \enquote{Modified distance calculation in generational distance and inverted generational distance,} \emph{International conference on evolutionary multi-criterion optimization}, Springer, 2015, pp. 110--125.

\bibitem[{David et~al.(2021)David, Delbecq, Defoort, Schmollgruber, Benard, and Pommier-Budinger}]{David_2021}
David, C., Delbecq, S., Defoort, S., Schmollgruber, P., Benard, E., and Pommier-Budinger, V., \enquote{From {FAST} to {FAST}-{OAD}: An open source framework for rapid Overall Aircraft Design,} \emph{{IOP} Conference Series: Materials Science and Engineering}, Vol. 1024, 2021, p. 012062.
\newblock \doi{10.1088/1757-899x/1024/1/012062}.

\bibitem[{Donelli et~al.(2022)Donelli, Ciampa, Lefebvre, Bartoli, Mello, Odaguil, and van~der Laan}]{dlr2022}
Donelli, G., Ciampa, P.~D., Lefebvre, T., Bartoli, N., Mello, J.~M., Odaguil, F.~I., and van~der Laan, T., \enquote{Value-driven Model-Based Optimization coupling Design-Manufacturing-Supply Chain in the Early Stages of Aircraft Development: Strategy and Preliminary Results,} \emph{AIAA AVIATION 2022 FORUM}, 2022.

\bibitem[{Saves et~al.(2022)Saves, Nguyen~Van, Bartoli, Lefebvre, David, Defoort, Diouane, and Morlier}]{SciTech_cat}
Saves, P., Nguyen~Van, E., Bartoli, N., Lefebvre, T., David, C., Defoort, S., Diouane, Y., and Morlier, J., \enquote{{Bayesian optimization for mixed variables using an adaptive dimension reduction process: applications to aircraft design},} \emph{{AIAA SciTech 2022}}, San Diego, United States, 2022.

\bibitem[{Saves et~al.(2021)Saves, Bartoli, Diouane, Lefebvre, Morlier, David, {Nguyen Van}, and Defoort}]{Aerobest_cat}
Saves, P., Bartoli, N., Diouane, Y., Lefebvre, T., Morlier, J., David, C., {Nguyen Van}, E., and Defoort, S., \enquote{Constrained bayesian optimization over mixed categorical variables, with application to aircraft design,} \emph{Proceedings of AeroBest 2021, ECCOMAS Thematic Conference on Multidisciplinary Design Optimization of Aerospace Systems}, 2021.

\end{thebibliography}

\end{document}